\documentclass[sigconf, nonacm]{acmart}
\renewcommand\footnotetextcopyrightpermission[1]{} 

\usepackage{booktabs}
\usepackage{multirow}
\usepackage{graphicx} 
\usepackage{subcaption} 
\usepackage{geometry} 
\usepackage{setspace} 
\usepackage[most]{tcolorbox}
\AtBeginDocument{%
  }

\setcopyright{none}
\copyrightyear{2018}
\acmYear{2018}
\acmDOI{XXXXXXX.XXXXXXX}
\acmConference[Conference acronym 'XX]{Make sure to enter the correct
  conference title from your rights confirmation email}{June 03--05,
  2018}{Woodstock, NY}
\acmISBN{978-1-4503-XXXX-X/2018/06}

\usepackage{titlesec}
\titlespacing*{\subsection}
  {0pt}{2pt}{1pt}
\titlespacing*{\section}
  {0pt}{3pt}{2pt}

\begin{document}
\captionsetup{font=footnotesize}
\captionsetup{skip=0pt} 
\setlength{\abovedisplayskip}{0pt} 
\setlength{\belowdisplayskip}{0pt} 
\setlength{\floatsep}{2pt}
\setlength{\textfloatsep}{2pt}
\setlength{\intextsep}{2pt}

\setlength{\dblfloatsep}{0pt} 
\setlength{\dbltextfloatsep}{0pt} 
\title{Revisiting Data Auditing in Large Vision-Language Models}


\author{Hongyu Zhu}
\affiliation{%
  \institution{Shanghai Jiao Tong University}
  \city{Shanghai}
  \country{China}}
\email{hongyu_z@sjtu.edu.cn}
\orcid{0009-0009-0570-8626}

\author{Sichu Liang}
\affiliation{%
  \institution{Southeast University}
  \city{Nanjing}
  \country{China}}
\email{coder_liang@seu.edu.cn}
\orcid{0009-0009-6798-1118}

\author{Wenwen Wang}
\affiliation{%
  \institution{Carnegie Mellon University}
  \city{Pittsburgh}
  \country{USA}}
\email{wenwenw@andrew.cmu.edu}
\orcid{0009-0009-9198-575X}

\author{Boheng Li}
\affiliation{%
  \institution{Nanyang Technological University}
  \city{Singapore}
  \country{Singapore}}
\email{BOHENG001@ntu.edu.sg}
\orcid{0000-0001-9921-7215}

\author{Tongxin Yuan}
\affiliation{%
  \institution{Shanghai Jiao Tong University}
  \city{Shanghai}
  \country{China}}
\email{teenyuan@sjtu.edu.cn}
\orcid{0009-0004-8167-8311}

\author{Fangqi Li}
\affiliation{%
  \institution{Shanghai Jiao Tong University}
  \city{Shanghai}
  \country{China}}
\email{solour_lfq@sjtu.edu.cn}
\orcid{0000-0001-7965-5170}

\author{ShiLin Wang}
\affiliation{%
  \institution{Shanghai Jiao Tong University}
  \city{Shanghai}
  \country{China}}
\email{wsl@sjtu.edu.cn}
\orcid{0000-0002-8214-6809}

\author{Zhuosheng Zhang}
\affiliation{%
  \institution{Shanghai Jiao Tong University}
  \city{Shanghai}
  \country{China}}
\email{zhangzs@sjtu.edu.cn}
\orcid{0000-0002-4183-3645}


\begin{abstract}
With the surge of large language models (LLMs), Large Vision-Language Models (VLMs)—which integrate vision encoders with LLMs for accurate visual grounding—have shown great potential in tasks like generalist agents and robotic control. However, VLMs are typically trained on massive web-scraped images, raising concerns over copyright infringement and privacy violations, and making data auditing increasingly urgent. Membership inference (MI), which determines whether a sample was used in training, has emerged as a key auditing technique, with promising results on open-source VLMs like LLaVA (AUC > 80\%). In this work, we revisit these advances and uncover a critical issue: current MI benchmarks suffer from distribution shifts between member and non-member images, introducing shortcut cues that inflate MI performance. We further analyze the nature of these shifts and propose a principled metric based on optimal transport to quantify the distribution discrepancy. To evaluate MI in realistic settings, we construct new benchmarks with i.i.d. member and non-member images. Existing MI methods fail under these unbiased conditions, performing only marginally better than chance. Further, we explore the theoretical upper bound of MI by probing the Bayes Optimality within the VLM’s embedding space and find the irreducible error rate remains high. Despite this pessimistic outlook, we analyze why MI for VLMs is particularly challenging and identify three practical scenarios—fine-tuning, access to ground-truth texts, and set-based inference—where auditing becomes feasible. Our study presents a systematic view of the limits and opportunities of MI for VLMs, providing guidance for future efforts in trustworthy data auditing.
\end{abstract}




\maketitle
\section{Introduction}
Large Vision-Language Models (VLMs) are becoming ubiquitous. Proprietary systems such as GPT-4o \cite{openai2024gpt4o} and Claude 3.5 Sonnet \cite{anthropic2024claude35sonnet} exhibit impressive multimodal capabilities, excelling at comprehensive image description and complex visual reasoning, with promising applications in generalist agents \cite{ge2024worldgpt} and embodied robotics \cite{mon2025embodied}. To support open scientific research, the community has made notable progress in replicating these abilities under transparent, open-source settings. Notably, the LLaVA series \cite{liu2023visual, li2025llava} integrates vision encoders with large language models (LLMs), achieving competitive performance while remaining fully accessible.

However, training large VLMs typically involves scraping web-scale multimodal data \cite{openai2024gpt4o, bai2025qwen2}, raising concerns about data legality and transparency. Recent incidents have highlighted these risks: copyright lawsuits involving GPT-4o \cite{kageyama2025ghibli}, potential personal data leakage via VLM outputs \cite{nasr2025scalable}, and test set contamination in benchmark evaluations \cite{oren2023proving, robison2025meta}. These challenges underscore the urgent need for principled data auditing, empowering third-party verification of \textit{whether specific data were used during VLM training}. 

Membership inference (MI)—which assesses whether a specific sample was part of a model’s training set \cite{shokri2017membership}—has recently emerged as a powerful approach for auditing VLMs. In this setting, MI probes a black-box API with a target image and a crafted instruction, analyzing the language response and token-level probabilities to compute membership-indicative statistics. Thresholds calibrated on known member and non-member images are then used to infer membership status \cite{li2024membership, hu2025membership}, enabling non-intrusive and statistically grounded audits. Recent work has introduced benchmark datasets for open-source VLMs like LLaVA, designating subsets of training images as members and post-release or synthetic images as non-members \cite{li2024membership}. Under this setup, state-of-the-art (SOTA) MI methods have achieved promising results (e.g., AUC > 80\%).

\begin{figure*}[hbtp] 
    \centering
    \includegraphics[width=0.8\linewidth]{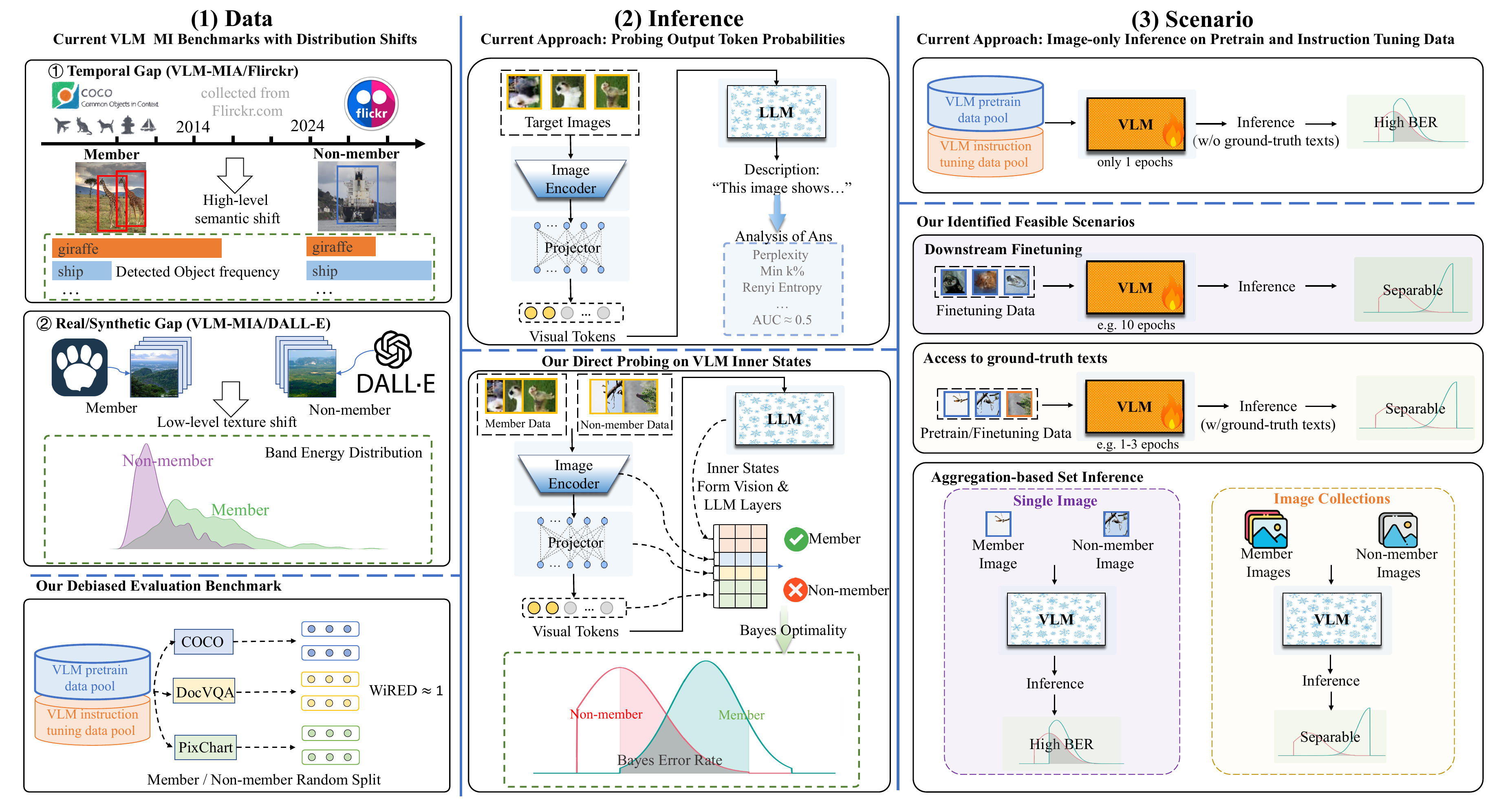} 
    \caption{Revisiting of VLM MI: (1) Identifying Bias in MI Datasets, (2) Probing Bayes Optimality in VLM Inner States, (3) Future Scenarios for Data Auditing.} 
    \label{fig:pipeline} 
\end{figure*}

However, this paper identifies a critical issue in current VLM MI benchmarks: \textbf{distribution shifts between member and non-member images introduce unintended shortcuts for inference}. These shifts stem from long-term temporal drift or discrepancies between real and synthetic image sources. We show that a simple image-only classifier (e.g., EfficientNet-B0 \cite{tan2019efficientnet}), outperforms most SOTA MI methods—without accessing any VLM outputs (Table~\ref{tab:f_d_large}). Moreover, when evaluating on \textit{pseudo-MI} datasets where all samples share the same membership status but differ in distribution, MI methods still perform well in distinguishing subsets, indicating that their success is driven largely by distributional artifacts rather than genuine membership signals (Table~\ref{tab:m_n_large}).

Thus, to ensure reliable inference, it is critical that member and non-member images are drawn from the same underlying distribution. Unfortunately, distribution shifts in the visual domain are pervasive yet often imperceptible \cite{geirhos2020shortcut}, arising in subtle or multifaceted ways \cite{zengunderstanding}. To support debiasing efforts, we systematically analyze concrete forms of shift by interpretably encoding images with a visual bag-of-words (capturing high-level semantics) and a frequency-domain energy profile (capturing low-level textures). To quantify these discrepancies, we further introduce WiRED—a principled metric that measures the ratio of sliced Wasserstein distances across embedding spaces tailed to different types of shift.

Building on this, we construct a suite of unbiased MI datasets by carefully examining four open-source VLM families—LLaVA-1.5 \cite{liu2024improved}, LLaVA-OneVision \cite{li2025llava}, Cambrian-1 \cite{tong2024cambrian}, and Molmo \cite{deitke2025molmo}. Leveraging random train/test splits from pretraining and instruction-tuning data, the member and non-member images are ensured to be drawn i.i.d. Under these conditions, SOTA MI methods perform only slightly better than random guessing. To assess the true auditing potential of MI, we design a series of classifiers (e.g., attention pooling probes) to directly inspect the VLM embedding space—where memorization signals are expected to reside. Even in this idealized setting, separability remains poor. We further estimate the \textit{theoretical upper bound} of inference accuracy via Bayes optimality and observe that the irreducible error remains substantial.

These sobering results raise a central question: why is MI particularly challenging for VLMs? Through careful analysis, we identify three key factors. First, the massive data volume and single-epoch training lead to minimal overfitting. Second, model developers relabel images with high-quality captions, rendering the ground-truth text inaccessible. Third, inherent attributes of a single image result in high-variance token confidences, diluting the membership signal. To address these challenges, we relax standard assumptions and propose three practical scenarios: multi-epoch fine-tuning on downstream tasks, access to ground-truth text, and aggregation-based set inference. Under all three settings, MI becomes not only feasible but also practically valuable for real-world data auditing.

Our analytical framework is illustrated in Figure~\ref{fig:pipeline}. The key contributions of this work are as follows:
\begin{itemize}
    \item We identify distribution shifts in existing VLM MI benchmarks, characterize their concrete forms, and introduce a principled metric WiRED to quantify them (§~\ref{sec:shortcut}).
    
    \item We construct unbiased MI datasets with i.i.d. splits, where SOTA MI methods perform marginally better than chance. We further estimate the Bayes optimality in VLM embedding space to assess the theoretical limits of MI (§~\ref{sec:feasibility}).
    
    \item We analyze why MI is particularly challenging for VLMs and identify three realistic scenarios in which MI becomes feasible and practically relevant for data auditing (§~\ref{sec:when}).
\end{itemize}

\section{Related Work}
\subsection{Large Vision-Language Models}
Built upon powerful LLMs~\cite{openai2022chatgpt}, proprietary VLMs such as GPT-4V~\cite{openai2023gpt4v} have shown impressive performance on open-domain multimodal tasks. Open-source efforts like LLaVA~\cite{liu2023visual} follow a simple design, aligning vision and language via a vision encoder $g_\psi(\cdot)$, a projector $p_\theta(\cdot)$, and a language model $f_\phi(\cdot)$. Given an image $X_v$ and instruction $X_q$, features $Z_v = g_\psi(X_v)$ are projected to visual tokens $H_v = p_\theta(Z_v)$ and input to the LLM with $X_q$ for response generation. Training involves modality alignment on large-scale datasets (e.g., LAION-5B~\cite{schuhmann2022laion}, CC12M~\cite{jia2021scaling}, Datacomp~\cite{gadre2023datacomp}) and instruction tuning on curated multimodal QA tasks.

To promote transparency, recent open-source VLMs such as LLaVA-1.5 \cite{liu2024improved}, LLaVA-OneVision \cite{li2025llava}, Cambrian-1 \cite{tong2024cambrian}, and Molmo \cite{deitke2025molmo} release not only model weights but also complete training data. As web-crawled image-text pairs are often noisy or short, these models re-annotate images with synthetic or human-curated texts. LLaVA-OneVision generates 99.8\% of its high-quality knowledge with proprietary VLMs~\cite{chen2024sharegpt4v}, while Molmo constructs its pretraining data from human-transcribed speech descriptions of web images.

\subsection{Membership Inference on Language Models}
As foundational models are trained on large-scale web data, concerns about unauthorized use of copyrighted or private content have intensified~\cite{kim2023propile, duarte2024cop}. Membership inference (MI) has emerged as a non-intrusive auditing tool to detect training data exposure~\cite{shokri2017membership}. Early MI methods relied on simple statistics like perplexity~\cite{carlini2021extracting, yeom2018privacy}, while recent work proposes token-level metrics like \textit{Min-K\%} \cite{shidetecting} and \textit{Min-K\%++} \cite{zhang2025min} for improved stability.
In the VLM setting, the standard threat model assumes black-box access, with only an image $X_v$ and crafted instruction $X_q$ available. Auditors collect output token probabilities and compute MI scores. VL-MIA~\cite{li2024membership} builds the first VLM MI benchmark and introduces \textit{MaxRényi-K\%}, while image-only inference~\cite{hu2025membership} assesses membership via self-consistency of sampled descriptions. As these methods produce only MI scores rather than binary decisions, evaluation relies on threshold-free metrics such as AUC and TPR@5\%FPR, while real-world deployment requires reference sets to calibrate decision thresholds.

Despite advances in \textit{LLM MI}, recent work has revealed temporal biases in evaluation datasets \cite{maini2024llm}, where non-members contain low-confidence, out-of-vocabulary tokens. In this work, we reveal a similar but subtler distribution shift in \textit{VLM MI}: member and non-member images exhibit complex, vision-specific biases that are difficult to detect or interpret. Unlike textual shifts explainable by out-of-vocabulary words \cite{das2024blind}, these visual biases are largely opaque. We introduce a principled framework to quantify such shifts and assess their impact on real-world data auditing.





\section{Distribution Shortcuts in VLM Membership Inference Benchmarks}
\label{sec:shortcut}
We identify that in current VLM MI benchmarks, distribution shifts between member and non-member images contribute significantly more to their separability than the genuine membership signal. Notably, on benchmarks such as VL-MIA/Flickr or VL-MIA/DALL-E, an EfficientNet-B0 \cite{tan2019efficientnet} trained solely on 300 images—without any access to VLM outputs—achieves a test AUC well above most SOTA MI methods. Moreover, MI methods yield high AUCs when comparing subsets with identical membership status yet different underlying distributions (\S~\ref{sec:distri_shift}). 

This pervasive distribution shift not only compromises evaluations but also limits the practical utility of MI. When target images deviate in distribution from the reference set used to calibrate decision thresholds, the derived criteria become misaligned, causing substantial false positives and false negatives. To address this, we provide a thorough analysis of the concrete form of distribution shifts in VL-MIA/Flickr and VL-MIA/DALL-E (\S~\ref{sec:understand}), and introduce a principled measure of distribution discrepancy that facilitates benchmark debiasing and reliable data auditing (\S~\ref{sec:quantify}).

\subsection{Experimental Settings}

\subsubsection{Datasets.}
Our study builds upon the existing VLM MI benchmark: VL-MIA/Flickr and VL-MIA/DALL-E, each comprising 300 member and 300 non-member images~\cite{li2024membership}.
In VL-MIA/Flickr, member images are sourced from MS COCO~\cite{lin2014microsoft}, a dataset widely used in training open-source VLMs, while non-member images are scraped from Flickr~\cite{flickr} after 2024. This temporal split ensures that member images were seen during VLM training, while non-members were not. However, since MS COCO was collected from Flickr prior to 2014, the two subsets are separated by over a decade. Even with identical collection pipelines, such a temporal gap naturally introduces distribution shifts due to evolving societal context, technological advancements, and user behavior~\cite{engstrom2020identifying}.
In VL-MIA/DALL-E, the member set is drawn from LAION~\cite{schuhmann2022laion}, while non-members are synthesized by DALL-E~\cite{dalle_openai} using the exact captions associated with member images. Although the two sets share identical textual descriptions, a distribution gap persists due to the inherent differences between real and synthesized images.
This raises a fundamental question: is the observed separability truly due to membership signals, or merely a result of distributional discrepancies? Our study aims to disentangle the two sources of separability.
\subsubsection{Vision-Language Models.}
We consider all representative VLM families with fully open data and models, enabling complete training data traceability, as listed in ~\cite{deitke2025molmo}. In this section, we evaluate LLaVA-1.5-7B~\cite{liu2023visual, liu2024improved}, LLaVA-OneVision-7B~\cite{li2025llava}, Cambrian-1-8B~\cite{tong2024cambrian}, and Molmo-7B-D~\cite{deitke2025molmo}.
\subsubsection{Membership Inference Methods.}  
We consider a range of MI methods, from classical approaches such as \textit{perplexity} and \textit{maximum probability gap} \cite{carlini2021extracting, yeom2018privacy}, to recent SOTA methods for LLMs, \textit{Min-$K\%$} \cite{shidetecting} (ICLR 2024) and \textit{Min-$K\%$++} \cite{zhang2025min} (ICLR 2025).
We also include VLM-specific methods: \textit{MaxRényi-$K\%$} and \textit{ModRényi} \cite{li2024membership} (NeurIPS 2024), as well as the \textit{Image-only Inference}~\cite{hu2025membership} (USENIX Security 2025).
All methods interact with VLMs via their language interface, simulating realistic black-box auditing: given the target image and a crafted instruction, the auditor infers membership from the VLM’s language response and its confidence scores.
\subsubsection{The Blind Classifier.}
To isolate the impact of distribution shift from genuine membership signals, we evaluate a \textit{blind} classifier without access to any VLM outputs. An EfficientNet-B0 \cite{tan2019efficientnet} is trained directly on images from VL-MIA/Flickr and VL-MIA/DALL-E to distinguish members from non-members, using a 1:1 split of 300 training and 300 testing samples per dataset.

\subsection{Distribution Shifts Surpass Membership Signals in Separability}
\label{sec:distri_shift}
\begin{table}[htbp]
  \centering
  \caption{Performance of MI methods on Existing MI Datasets.}
  \resizebox{\linewidth}{!}{
    \begin{tabular}{ccc|cccccccc}
    \toprule[1.5pt]
    \multirow{2}[0]{*}{\textbf{Dataset}} & \multicolumn{2}{c|}{\multirow{2}[0]{*}{\textbf{Method}}} & \multicolumn{2}{c}{\textbf{LLaVA-1.5}} & \multicolumn{2}{c}{\textbf{LLaVA-ov}} & \multicolumn{2}{c}{\textbf{Molmo}} & \multicolumn{2}{c}{\textbf{Cambrian}} \\
    \cmidrule(lr){4-5} \cmidrule(lr){6-7} \cmidrule(lr){8-9} \cmidrule(lr){10-11}
          & \multicolumn{2}{c|}{} & \textbf{AUC}   & \textbf{TPR}   & \textbf{AUC}   & \textbf{TPR}   & \textbf{AUC}   & \textbf{TPR}   & \textbf{AUC}   & \textbf{TPR} \\
          \midrule
    \multirow{14}[0]{*}{\textbf{\shortstack{Flickr\\(WiRED\\=2.10)}}} & \multirow{2}[0]{*}{\textbf{Perplexity}} & \textbf{inst} & 34.3  & 0.7   & 18.1  & 0.0   & 34.6  & 1.3   & 34.4  & 1.7  \\
          &       & \textbf{desp} & 57.4  & \underline{16.0}  & 14.5  & 0.0   & 55.8  & 12.0  & 43.3  & 4.0  \\
          & \multirow{2}[0]{*}{\textbf{Max-Prob-Gap}} & \textbf{inst} & 57.5  & 7.3   & 21.8  & 0.7   & \underline{67.3}  & 11.0  & 9.4   & 0.3  \\
          &       & \textbf{desp} & 61.0  & \underline{16.7}  & 62.1  & 9.3   & 56.8  & 10.3  & 36.7  & 1.3  \\
          & \multirow{2}[0]{*}{\textbf{Min-K\%}} & \textbf{inst} & 48.2  & 1.7   & 59.9  & 10.0  & \underline{65.1}  & 12.7  & 55.1  & 4.0  \\
          &       & \textbf{desp} & 57.3  & \underline{16.0}  & 20.1  & 0.0   & 55.8  & 12.3  & 59.4  & 12.0  \\
          & \multirow{2}[0]{*}{\textbf{Min-K\%++}} & \textbf{inst} & 54.5  & 13.7  & 14.7  & 0.3   & \underline{65.5}  & 13.3  & 44.7  & 1.7  \\
          &       & \textbf{desp} & 61.8  & \underline{21.7}  & \underline{80.1}  & 14.0  & \underline{67.9}  & \underline{16.3}  & 50.7  & 1.0  \\
          & \multirow{2}[0]{*}{\textbf{MaxRényi-K\% }} & \textbf{inst} & \underline{66.6}  & \underline{18.7}  & 48.3  & 8.3   & \underline{69.3}  & \underline{18.7}  & \underline{68.4}  & \underline{16.3}  \\
          &       & \textbf{desp} & 57.4  & \underline{19.3}  & \underline{95.5}  & \underline{79.0}  & 64.3  & \underline{19.0}  & 55.5  & 7.7  \\
          & \multirow{2}[0]{*}{\textbf{ModRényi}} & \textbf{inst} & 38.4  & 0.3   & \underline{84.2}  & \underline{58.7}  & 32.7  & 0.7   & 41.4  & 1.3  \\
          &       & \textbf{desp} & 57.1  & 12.3  & 15.7  & 0.0   & 53.0  & 10.0  & 45.5  & 3.7  \\
          & \multicolumn{2}{c|}{\textbf{Image-only Inference}} & 64.0  & 8.7   & 21.6  & 6.5   & \underline{66.8}  & 13.5  & \underline{76.1}  & \underline{31.6}  \\
          & \multicolumn{2}{c|}{\textbf{Blind Classifier (ours)}} & \underline{\textbf{99.1 }} & \underline{\textbf{97.4 }} & \underline{\textbf{99.1 }} & \underline{\textbf{97.4 }} & \underline{\textbf{99.1 }} & \underline{\textbf{97.4 }} & \underline{\textbf{99.1 }} & \underline{\textbf{97.4 }} \\
    \midrule
    \multirow{14}[0]{*}{\textbf{\shortstack{DALL-E\\(WiRED\\=3.00)}}} & \multirow{2}[0]{*}{\textbf{Perplexity}} & \textbf{inst} & 37.7  & 2.0   & 58.4  & \underline{17.0}  & 59.0  & 11.3  & 24.6  & 1.0  \\
          &       & \textbf{desp} & \underline{65.5}  & 8.0   & 55.5  & \underline{17.3}  & 55.9  & 12.0  & \underline{71.2}  & 12.7  \\
          & \multirow{2}[0]{*}{\textbf{Max-Prob-Gap}} & \textbf{inst} & 58.9  & 9.3   & 63.3  & \underline{16.3}  & \underline{67.8}  & \underline{15.3}  & \underline{89.3}  & \underline{66.3}  \\
          &       & \textbf{desp} & 64.8  & 7.7   & 58.0  & \underline{17.3}  & 58.4  & 14.7  & \underline{78.5}  & \underline{38.3}  \\
          & \multirow{2}[0]{*}{\textbf{Min-K\%}} & \textbf{inst} & 39.3  & 6.3   & 60.5  & \underline{18.7}  & 63.5  & 11.3  & 33.9  & 1.0  \\
          &       & \textbf{desp} & \underline{65.6}  & 8.0   & \underline{65.7}  & \underline{18.0}  & 55.9  & 12.0  & \underline{70.8}  & 12.0  \\
          & \multirow{2}[0]{*}{\textbf{Min-K\%++}} & \textbf{inst} & 58.5  & 10.3  & 54.7  & 12.7  & 48.2  & 7.0   & \underline{87.7}  & \underline{66.7}  \\
          &       & \textbf{desp} & \underline{67.0}  & 9.3   & 49.6  & 10.7  & 57.6  & 11.0  & \underline{75.4}  & 14.7  \\
          & \multirow{2}[0]{*}{\textbf{MaxRényi-K\% }} & \textbf{inst} & \underline{71.8}  & 14.3  & \underline{78.7}  & \underline{30.3}  & 54.5  & 7.7   & \underline{\textbf{94.9}}  & \underline{\textbf{87.7}}  \\
          &       & \textbf{desp} & \underline{70.3}  & 9.0   & 64.0  & 14.3  & 57.9  & 14.7  & \underline{87.1}  & \underline{56.3}  \\
          & \multirow{2}[0]{*}{\textbf{ModRényi}} & \textbf{inst} & 38.4  & 3.0   & \underline{71.9}  & \underline{29.0}  & 51.0  & 9.3   & 25.5  & 5.0  \\
          &       & \textbf{desp} & 64.6  & 9.0   & 62.9  & \underline{24.3}  & 55.0  & 12.3  & \underline{84.0}  & \underline{47.3}  \\
& \multicolumn{2}{c|}{\textbf{Image-only Inference}} & 47.0  & 4.9   & \underline{67.3}  & 14.2  & 63.7  & 13.7  & 41.0  & 0.0  \\

          & \multicolumn{2}{c|}{\textbf{Blind Classifier (ours)}} & \underline{\textbf{87.4 }} & \underline{\textbf{49.1 }} & \underline{\textbf{87.4 }} & \underline{\textbf{49.1 }} & \underline{\textbf{87.4 }} & \underline{\textbf{49.1 }} & \underline{87.4}  & \underline{49.1}  \\
          \bottomrule[1.5pt]
    \end{tabular}%
    }
  \label{tab:f_d_large}%
\end{table}%

Table~\ref{tab:f_d_large} reports the performance of MI methods on VL-MIA/Flickr and VL-MIA/DALL-E. Following standard practice \cite{shidetecting, zhang2025min, li2024membership, hu2025membership}, the Area Under the ROC Curve (AUC) and True Positive Rate at 5\% False Positive Rate (TPR@5\%FPR) are used as evaluation protocols, where higher values indicate better inference. Surprisingly, most of the times, the EfficientNet-B0 trained solely on images achieves significantly higher scores than SOTA MI methods, indicating that distribution shifts between member and non-member images alone are sufficient for separation, independent of any memorization signals from the VLM.

However, could the supervised training simply encourage EfficientNet to exploit distributional cues, while SOTA MI methods genuinely reflect the VLM’s overfitting behavior? To disentangle the effects of distribution shift from true membership signals, we construct two \textit{pseudo-MI} datasets by swapping subsets between VL-MIA/Flickr and VL-MIA/DALL-E: 
in VL-MIA/Member, all images come from the training set—VL-MIA/Flickr members as members, and VL-MIA/DALL-E members as \textit{pseudo-nonmembers}. Similarly, in VL-MIA/Nonmember, all images are unseen during training, with VL-MIA/DALL-E nonmembers as nonmembers, and VL-MIA/Flickr nonmembers as \textit{pseudo-members}. While membership status is uniform within each dataset, distributional discrepancies remain.

\begin{table}[htbp]
  \centering
  \caption{Performance of MI methods on \textit{Pseudo-MI} Datasets}
  \resizebox{\linewidth}{!}{
    \begin{tabular}{ccc|cccccccc}
    \toprule[1.5pt]
    \multirow{2}[0]{*}{\textbf{Dataset}} & \multicolumn{2}{c|}{\multirow{2}[0]{*}{\textbf{Method}}} & \multicolumn{2}{c}{\textbf{LLaVA-1.5}} & \multicolumn{2}{c}{\textbf{LLaVA-ov}} & \multicolumn{2}{c}{\textbf{Molmo}} & \multicolumn{2}{c}{\textbf{Cambrian}} \\
    \cmidrule(lr){4-5} \cmidrule(lr){6-7} \cmidrule(lr){8-9} \cmidrule(lr){10-11}
          & \multicolumn{2}{c|}{} & \textbf{AUC}   & \textbf{TPR}   & \textbf{AUC}   & \textbf{TPR}   & \textbf{AUC}   & \textbf{TPR}   & \textbf{AUC}   & \textbf{TPR} \\
            \midrule
    \multirow{14}[0]{*}{\textbf{\shortstack{Member\\(WiRED\\=2.42)}}} & \multirow{2}[0]{*}{\textbf{Perplexity}} & \textbf{inst} & 42.4  & 0.7   & 7.9   & 0.0   & 20.4  & 0.0   & 54.5  & 1.0  \\
          &       & \textbf{desp} & 63.9  & 7.0   & 20.4  & 0.0   & 51.9  & 2.0   & 27.2  & 1.3  \\
          & \multirow{2}[0]{*}{\textbf{Max-Prob-Gap}} & \textbf{inst} & 53.8  & 0.3   & 4.0   & 0.0   & 37.0  & 0.3   & 1.7   & 0.0  \\
          &       & \textbf{desp} & 58.9  & 9.3   & 60.1  & 1.0   & 46.7  & 0.7   & 14.9  & 0.0  \\
          & \multirow{2}[0]{*}{\textbf{Min-K\%}} & \textbf{inst} & 42.6  & 1.0   & 48.2  & 0.0   & 39.1  & 0.7   & \underline{65.1}  & 6.3  \\
          &       & \textbf{desp} & \underline{65.2}  & 7.0   & 23.7  & 0.0   & 55.6  & 6.0   & \underline{76.8}  & \underline{21.7}  \\
          & \multirow{2}[0]{*}{\textbf{Min-K\%++}} & \textbf{inst} & \underline{76.0}  & \underline{16.0}  & 31.1  & 0.0   & \underline{77.5}  & 7.7   & 17.4  & 0.0  \\
          &       & \textbf{desp} & 65.2  & \underline{19.7}  & \underline{81.7}  & \underline{30.0}  & 45.2  & 3.7   & 29.5  & 0.0  \\
          & \multirow{2}[0]{*}{\textbf{MaxRényi-K\% }} & \textbf{inst} & \underline{84.8}  & \underline{38.0}  & \underline{66.0}  & 11.0  & \underline{82.0}  & \underline{29.7}  & 54.8  & 3.0  \\
          &       & \textbf{desp} & \underline{78.7}  & \underline{18.0}  & \underline{90.9}  & \underline{52.3}  & 55.4  & 5.3   & 45.4  & 4.7  \\
          & \multirow{2}[0]{*}{\textbf{ModRényi}} & \textbf{inst} & 38.6  & 0.0   & \underline{84.0}  & \underline{40.3}  & 18.4  & 0.0   & \underline{65.8}  & 2.7  \\
          &       & \textbf{desp} & 63.9  & 7.3   & 18.7  & 0.0   & 53.4  & 3.0   & 37.3  & 1.3  \\
& \multicolumn{2}{c|}{\textbf{Image-only Inference}} & \underline{68.2}  & \underline{19.7}  & 20.0  & 4.4   & 41.6  & 2.0   & 59.9  & 14.9  \\

          & \multicolumn{2}{c|}{\textbf{Blind Classifier (ours)}} & \underline{\textbf{97.3 }} & \underline{\textbf{85.5 }} & \underline{\textbf{97.3 }} & \underline{\textbf{85.5 }} & \underline{\textbf{97.3 }} & \underline{\textbf{85.5 }} & \underline{\textbf{97.3 }} & \underline{\textbf{85.5 }} \\

    \midrule
    \multirow{14}[0]{*}{\textbf{\shortstack{Non\\Member\\(WiRED\\=6.60)}}} & \multirow{2}[0]{*}{\textbf{Perplexity}} & \textbf{inst} & 35.0  & 1.3   & 34.9  & 4.3   & 41.7  & 0.3   & 37.0  & 0.0  \\
          &       & \textbf{desp} & 57.2  & 3.7   & \underline{67.7}  & \underline{33.3}  & 52.9  & 3.3   & 57.1  & 4.3  \\
          & \multirow{2}[0]{*}{\textbf{Max-Prob-Gap}} & \textbf{inst} & 55.0  & 2.7   & 22.9  & 0.3   & 41.8  & 1.3   & 50.8  & 7.7  \\
          &       & \textbf{desp} & 55.5  & 2.3   & 56.9  & 10.0  & 48.8  & 4.0   & 54.0  & 6.0  \\
          & \multirow{2}[0]{*}{\textbf{Min-K\%}} & \textbf{inst} & 45.2  & 6.7   & \underline{65.7}  & \underline{22.0}  & 46.2  & 0.3   & 44.3  & 1.3  \\
          &       & \textbf{desp} & 57.3  & 4.3   & \underline{72.2}  & \underline{34.0}  & 60.1  & 8.7   & 57.0  & 8.0  \\
          & \multirow{2}[0]{*}{\textbf{Min-K\%++}} & \textbf{inst} & \underline{75.0}  & \underline{18.0}  & \underline{75.6}  & \underline{18.7}  & 63.6  & 6.0   & \underline{66.4}  & \underline{16.7}  \\
          &       & \textbf{desp} & 59.8  & 6.3   & 44.6  & 10.0  & 34.6  & 2.3   & 51.6  & 8.0  \\
          & \multirow{2}[0]{*}{\textbf{MaxRényi-K\% }} & \textbf{inst} & \underline{79.9}  & \underline{21.3}  & \underline{90.5}  & \underline{63.3}  & \underline{81.3}  & \underline{20.7}  & \underline{84.4}  & \underline{47.7}  \\
          &       & \textbf{desp} & \underline{66.9}  & 13.3  & \underline{79.9}  & \underline{36.3}  & 49.6  & 4.0   & 55.9  & 7.7  \\
          & \multirow{2}[0]{*}{\textbf{ModRényi}} & \textbf{inst} & 35.2  & 0.7   & 44.4  & 4.7   & 43.4  & 0.0   & 36.1  & 3.7  \\
          &       & \textbf{desp} & 57.5  & 4.7   & \underline{73.7}  & \underline{38.3}  & 57.5  & 5.0   & 58.2  & \underline{13.0}  \\
& \multicolumn{2}{c|}{\textbf{Image-only Inference}} & 48.6  & 8.1   & 58.2  & 5.8   & 40.3  & 1.8   & 31.4  & 0.0  \\
                    & \multicolumn{2}{c|}{\textbf{Blind Classifier (ours)}} & \underline{\textbf{99.6 }} & \underline{\textbf{98.4 }} & \underline{\textbf{99.6 }} & \underline{\textbf{98.4 }} & \underline{\textbf{99.6 }} & \underline{\textbf{98.4 }} & \underline{\textbf{99.6 }} & \underline{\textbf{98.4 }} \\
          \bottomrule[1.5pt]
    \end{tabular}%
    }
  \label{tab:m_n_large}%
\end{table}%

As shown in Table~\ref{tab:m_n_large}, despite the absence of genuine membership differences, MI methods retain high performance on these \textit{pseudo-MI} datasets \footnote{AUC > 65\% and TPR > 15\% are highlighted with underlines.}, matching or exceeding their performance on the original benchmarks. The blind EfficientNet continues to perform best, confirming that separability primarily stems from distribution shifts. 
Notably, some methods yield AUCs well below 50\% (random guessing), despite balanced datasets. This reflects the assumption in MI methods that membership signals are directional—e.g., lower \textit{perplexity} implies membership. However, distribution shifts can invalidate this assumption, resulting in inverted decisions. In general, AUCs near 50\% indicate low separability.

\begin{tcolorbox}[colback=gray!5!white,
                  colframe=teal!80!black,
                  coltitle=black,
                  fonttitle=\bfseries\itshape,
                  enhanced,
                  sharp corners=southwest,
                  drop shadow,
                after=\vspace{0pt}
                  ]
Finding 1: Current VLM MI benchmarks exhibit distribution shifts between member and non-member images, acting as unintended shortcuts. SOTA MI methods rely on these shifts rather than genuine membership signals.
\end{tcolorbox}

\subsection{Understanding Distribution Discrepancy}
\label{sec:understand}
The strong performance of a blind EfficientNet indicates the presence of distribution shifts. Yet, the exact nature of these shifts remains unclear: what specific discrepancies are being exploited by MI methods?
Recent studies~\cite{liu2025decade} show that shifts are common in large-scale datasets like LAION~\cite{schuhmann2022laion} and DataComp~\cite{gadre2023datacomp}, but often imperceptible to humans~\cite{liu2025decade}. Auditors may be unaware that the target and reference images differ in distribution, undermining threshold calibration. Understanding the concrete forms of shifts is therefore critical for debiasing. To this end, we analyze VL-MIA/Flickr and VL-MIA/DALL-E from two perspectives: high-level semantics and low-level textures.

\begin{figure}[hbtp] 
    \centering
    \includegraphics[width=\linewidth]{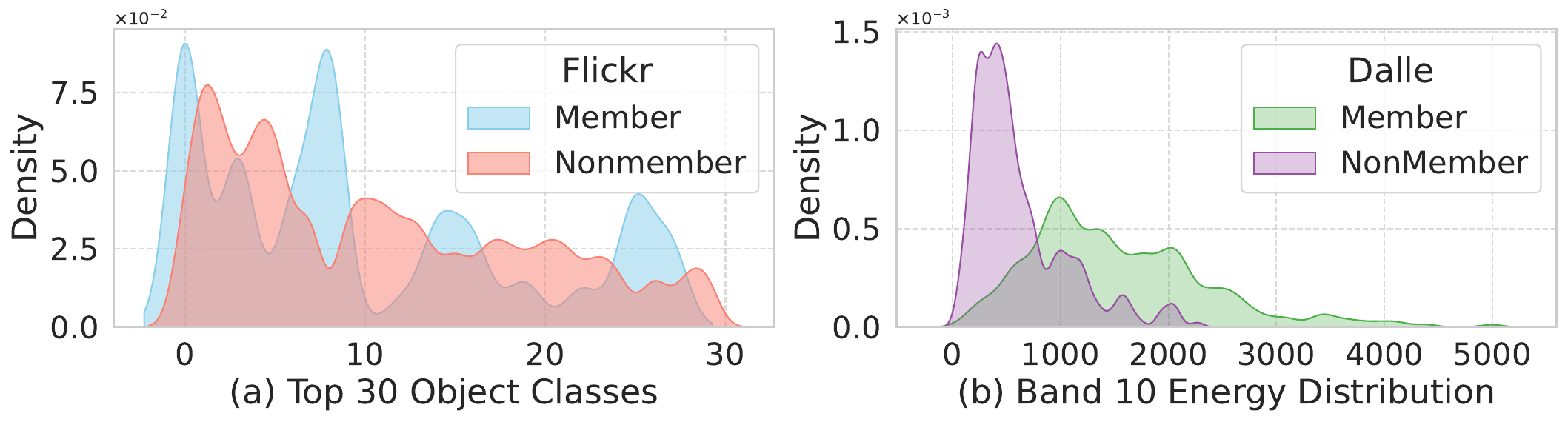} 
    \caption{Distribution Shifts in Existing MI Datasets: (a) Flickr; (b) DALL-E.} 
    \label{fig:obj_dist} 
\end{figure}

\textbf{VL-MIA/Flickr} exhibits a temporal gap of over a decade between member and non-member images, leading to semantic shifts in object content. Member images from MS COCO tend to depict natural scenes and wildlife (e.g., giraffes) with a more concentrated object distribution, while non-members—randomly crawled from Flickr after 2024—contain more man-made objects (e.g., ships) and a flatter distribution.
To validate this, we train a YOLOv11 detector \cite{ultralytics2024yolov11} on the LVIS dataset~\cite{gupta2019lvis} (1,203 categories) and extract object annotations from both subsets. As shown in Figure~\ref{fig:obj_dist}(a), the category frequency distributions differ significantly.
We further encode each image as a 1,203-dimensional sparse vector of object counts and train a random forest \cite{breiman2001random} to distinguish members from non-members. This visual bag-of-words classifier achieves an average test AUC of 81.96\%—outperforming most SOTA MI methods—demonstrating that high-level semantic shifts alone are sufficient to separate the two subsets in VL-MIA/Flickr.

In contrast, \textbf{VL-MIA/DALL-E} controls high-level semantics by generating non-member images with DALL-E using the same captions as member images, resulting in minimal semantic variation. Thus, the visual bag-of-words classifier performs poorly, with an AUC of 53.15\%—near random guessing.
However, real and AI-generated images often differ significantly in low-level details~\cite{lu2023seeing}. To explore this, we analyze high-frequency features, which are known to capture fine-grained visual cues~\cite{zengunderstanding}.
Given an image with pixel intensity matrix \( I \in \mathbb{R}^{H \times W} \), the centered magnitude spectrum of its 2D discrete Fourier transform is computed as:

\begin{equation} \mathcal{F}(u, v) = \sum_{x=0}^{H-1} \sum_{y=0}^{W-1} I(x, y) \cdot e^{-2\pi i \left(\frac{ux}{H} + \frac{vy}{W}\right)}, \quad M(u, v) = \left| \mathcal{F}(u, v) \right| \label{eq:dft} \end{equation}

We divide the spectrum into \(K\) concentric frequency bands based on \(\ell_2\) distance from the center. The energy of the \(i\)-th band is:
\begin{equation}
E_i = \frac{1}{|\mathcal{B}_i|} \sum_{(u,v) \in \mathcal{B}_i} M(u, v)
\label{eq:band_energy}
\end{equation}
where \(\mathcal{B}_i\) is the set of frequency coordinates in the \(i\)-th band. This defines a frequency feature vector \(\mathbf{E} = [E_0, E_1, \dots, E_{K-1}] \in \mathbb{R}^K\).  
Figure~\ref{fig:obj_dist}(b) shows the distribution of high-frequency energy \(E_{10}\) for \(K = 10\), revealing clear separation between members and non-members.  
Using \(\mathbf{E}\) as a 10-dimensional input, a simple linear classifier achieves a test AUC of 99.64\%—far exceeding all MI baselines and even the supervised EfficientNet (AUC \(\approx 80\%\)).  
Despite similar high-level semantics, subtle texture differences are sufficient to separate member and non-member images in VL-MIA/DALL-E.

\begin{tcolorbox}[colback=gray!5!white,
                  colframe=teal!80!black,
                  coltitle=black,
                  fonttitle=\bfseries\itshape,
                  enhanced,
                  sharp corners=southwest,
                  drop shadow,
                  after=\vspace{0pt}]
Finding 2: Distribution shifts are ubiquitous yet hard to detect. They can arise from temporal gaps or differences in data collection, and manifest in diverse forms—from subtle texture patterns to high-level semantic variations.
\end{tcolorbox}

\subsection{Quantifying Distribution Discrepancy}
\label{sec:quantify}
Accurately quantifying distribution shift is essential for building fair MI benchmarks and thresholds calibration. While shifts are pervasive in large-scale datasets, effective and interpretable metrics remain underexplored~\cite{liu2025decade, zengunderstanding}.
One straightforward approach is to train a deep classifier (e.g., EfficientNet) to distinguish subsets—an instantiation of classifier two-sample tests~\cite{lopez2017revisiting}. However, this is computationally expensive for personal privacy auditing and often underestimates subtle shifts, e.g., in VL-MIA/DALL-E, EfficientNet yields a significantly lower AUC than a frequency-based linear classifier.

Thus, to facilitate fair MI benchmarks and practical data auditing for VLMs, we propose a principled, interpretable, and efficient metric for quantifying distributional discrepancies. We introduce \textbf{WiRED}---\underline{W}asserste\underline{i}n \underline{R}atio of \underline{E}mbedde\underline{d} Representations---which measures the degree of shift between two image subsets \( S_1 \) and \( S_2 \).
Specifically, WiRED first embeds each image \( I \) into a collection of metric spaces via embedding functions \( \phi_1, \dots, \phi_t \), each targeting a distinct form of shift. Let \( p_1 \) and \( p_2 \) denote the probability densities of \( S_1 \) and \( S_2 \) in the \( i \)-th embedding space. The Wasserstein distance \cite{montesuma2024recent} between them is defined as:

\begin{equation}
W_q(p_1, p_2) = \left( \inf_{\gamma \in \Gamma(p_1, p_2)} \mathbb{E}_{(x_1, x_2) \sim \gamma} \| x_1 - x_2 \|^q \right)^{1/q},
\label{eq:wasserstein}
\end{equation}

where \( \Gamma(p_1, p_2) \) denotes all couplings between \( p_1 \) and \( p_2 \). Intuitively, \( W_q \) captures the minimal cost of transporting one distribution into the other, commonly referred to as the Earth Mover’s Distance \cite{montesuma2024recent}.
As computing \( W_q \) exactly requires solving the optimal transport problem with time complexity \( \mathcal{O}(N^3) \) for sample size \( N \), we adopt the sliced Wasserstein distance (SWD) \cite{bischoffpractical} as an efficient approximation, which projects samples onto random directions \( \{\theta_j\}_{j=1}^{K} \in \mathbb{R}^d \), computes one-dimensional Wasserstein distances, and averages them across all directions:

\begin{equation}
\text{SWD}(S_1, S_2) = \frac{1}{K} \sum_{j=1}^{K} W_q\left(\theta_j^\top \phi_i(S_1), \theta_j^\top \phi_i(S_2)\right).
\label{eq:swd}
\end{equation}

To normalize the discrepancy between \( S_1 \) and \( S_2 \), we compare their distance to the internal variation within \( S_1 \). Specifically, we sample two disjoint subsets \( S_{11}, S_{12} \subset S_1 \), along with a size-matched subset \( S_2' \subset S_2 \). The WiRED score in the \( i \)-th embedding space is then defined as:

\begin{equation}
\text{WiRED}_i = \frac{\text{SWD}(S_{11}, S_2')}{\text{SWD}(S_{11}, S_{12})}.
\label{eq:wired_i}
\end{equation}

This highlights how distinguishable \( S_2 \) is from \( S_1 \), relative to \( S_1 \)'s internal variation. A ratio close to 1 indicates similar distributions, while a significantly higher value signals a notable shift.
Since each embedding function \( \phi_i \) captures different aspects of the data, we define the final WiRED score as the maximum across all embeddings, i.e., \( \text{WiRED} = \max_{i \in [t]} \text{WiRED}_i \).

In our experiments, we instantiate \( \phi_i \) with two embedding functions: (1) ImageNet-pretrained EfficientNet-B0 features to capture high-level semantics, and (2) frequency-domain energy vectors (\S~\ref{sec:understand}) to capture low-level textures. WiRED is non-parametric, makes no assumptions about distribution shapes, and is highly efficient (e.g., taking ~10 seconds for a 600-image MI dataset). We report WiRED scores alongside each benchmark, clearly reflecting the distributional biases identified in \S~\ref{sec:distri_shift} (e.g., WiRED $\gg$ 1), without model training or prior knowledge of the shift. 

\section{Feasibility of Membership Inference on VLMs}
\label{sec:feasibility}
In \S~\ref{sec:shortcut}, we show that current VLM MI benchmarks suffer from distribution shifts, introducing unintended prediction shortcuts. When such shifts are eliminated, can MI reliably detect membership based on overfitting signals in VLM outputs?

In this section, we investigate this question and uncover that, under strictly i.i.d. conditions, SOTA MI methods perform only slightly better than random guessing (\S~\ref{sec:unbias}). Even with white-box access to VLM internal features—the presumed source of memorization—separability remains poor, and the theoretical upper bound of performance, quantified by the estimated Bayes Error Rate (BER), remains pessimistic (\S~\ref{sec:probe}). These results suggest that in the most realistic auditing scenario—where the auditor must determine whether a single image appeared in VLM training—current MI techniques are unlikely to yield reliable conclusions.

\subsection{Towards Unbiased VLM MI Datasets}
\label{sec:towards_dataset}
The most rigorous and straightforward way to construct i.i.d. subsets is through random splits from the same data source \cite{engstrom2020identifying}. Following this principle, we carefully inspect open-source VLMs to identify datasets with standard training/testing splits.
We focus on the four fully open VLM families—LLaVA-1.5 \cite{liu2024improved}, LLaVA OneVision \cite{li2025llava}, Cambrian-1 \cite{tong2024cambrian}, and Molmo \cite{deitke2025molmo}—all trained exclusively on publicly available data. Unlike smaller task-specific models, VLMs first align vision and language representations during pretraining, and are further tuned for instruction following. These phases typically utilize all available data, without held-out validation or test sets. Fortunately, many VLMs incorporate widely used captioning and VQA datasets that do provide standard splits. For instance, all four families use MS COCO \cite{lin2014microsoft}, primarily during pretraining, and both LLaVA OneVision and Molmo include instruction-tuning datasets built on clearly partitioned VQA benchmarks.

\begin{table}[htbp]
  \centering
  \caption{Quantitative Debiasing Validation: Blind Clf Performance and WiRED.}
  \resizebox{\linewidth}{!}{
    \begin{tabular}{cc|ccc}
    \toprule[1.5pt]
    \textbf{datasets} & \textbf{Models} & \textbf{blind clf AUC} & \textbf{blind clf TPR} & \textbf{WiRED} \\
    \midrule
    \textbf{COCO} & All models & 49.0±0.6 & 8.3±1.0 & 0.97  \\
    \cmidrule(lr){1-2}
    \textbf{ChartQA} & \multirow{3}[0]{*}{LLaVA-ov} & 57.7±0.7 & 8.3±1.6 & 1.04  \\
    \textbf{DocVQA} &       & 54.9±1.0 & 6.2±2.5 & 0.98  \\
    \textbf{InfoVQA} &       & 57.0±1.4 & 7.2±0.9 & 1.25  \\
    \cmidrule(lr){1-2}
    \textbf{PixMoChart} & \multirow{3}[0]{*}{Molmo} & 48.4±0.1 & 1.7±0.6 & 1.00  \\
    \textbf{PixMoDiagram} &       & 48.5±0.3 & 6.8±1.2 & 1.22  \\
    \textbf{PixMoTable} &       & 57.1±1.2 & 9.8±0.3 & 0.94  \\
    \bottomrule[1.5pt]
    \end{tabular}%
    }
  \label{tab:dataset}%
\end{table}%

To ensure that member images were seen during training while non-members were not, we select only datasets explicitly used in both training and evaluation, excluding any with evident distribution shifts between splits. The resulting datasets are listed in Table~\ref{tab:dataset}.
Following the VL-MIA setup \cite{li2024membership}, we sample 300 images each from the training and testing splits to form the member and non-member sets. We apply both EfficientNet and our proposed WiRED metric to assess distribution shifts. Across all selected datasets, EfficientNet classifiers yield AUCs near 50\%, and WiRED scores remain close to 1, confirming well-matched distributions. ChartQA \cite{masry2022chartqa}, DocVQA \cite{mathew2021docvqa}, and InfoVQA \cite{mathew2021docvqa}, show slightly elevated AUCs, though they adhere to standard train/test splits. We attribute this to image redundancy in the original VQA datasets; our strict de-duplication may introduce minor residual imbalance.
This effort represents an initial step toward reliable VLM MI benchmarks, paving the way for larger and more rigorous datasets in future research.

\subsection{VLM MI Performance on Unbiased Datasets}
\label{sec:unbias}
\begin{table}[htbp]
  \centering
  \caption{Performance of MI Methods on the Debiased COCO Dataset.}
  \resizebox{\linewidth}{!}{
    \begin{tabular}{cc|cccccccc}
        \toprule[1.5pt]
    \multicolumn{2}{c|}{\multirow{2}[0]{*}{\textbf{Method}}} & \multicolumn{2}{c}{\textbf{LLaVA-1.5}} & \multicolumn{2}{c}{\textbf{LLaVA-ov}} & \multicolumn{2}{c}{\textbf{Molmo}} & \multicolumn{2}{c}{\textbf{Cambrian}} \\
    \cmidrule(lr){3-4} \cmidrule(lr){5-6} \cmidrule(lr){7-8} \cmidrule(lr){9-10}
    \multicolumn{2}{c|}{} & \textbf{AUC} & \textbf{TPR} & \textbf{AUC} & \textbf{TPR} & \textbf{AUC} & \textbf{TPR} & \textbf{AUC} & \textbf{TPR} \\
    \midrule
    \multirow{2}[0]{*}{\textbf{Perplexity}} & \textbf{inst} & 51.2  & 6.3   & 50.7  & 7.0   & 49.3  & 5.7   & 55.3  & 8.3  \\
          & \textbf{desp} & 53.7  & 1.7   & 51.5  & 8.0   & 54.3  & 5.3   & 54.3  & 5.7  \\
    \multirow{2}[0]{*}{\textbf{Max-Prob-Gap}} & \textbf{inst} & 50.0  & 6.0   & 53.5  & 8.3   & 45.5  & 3.0   & 50.6  & 5.3  \\
          & \textbf{desp} & 53.0  & 5.7   & 53.5  & 4.3   & 51.8  & 8.3   & 54.5  & 8.0  \\
    \multirow{2}[0]{*}{\textbf{Min-K\%}} & \textbf{inst} & 51.2  & 7.3   & 50.7  & 7.0   & 51.5  & 5.7   & 55.8  & 8.7  \\
          & \textbf{desp} & 54.2  & 3.3   & 52.6  & 8.0   & 54.8  & 7.0   & 54.1  & 7.3  \\
    \multirow{2}[0]{*}{\textbf{Min-K\%++}} & \textbf{inst} & 50.3  & 4.3   & 52.4  & 6.3   & 51.6  & 6.3   & 52.5  & 7.3  \\
          & \textbf{desp} & 53.4  & 8.3   & 52.9  & 8.7   & 51.0  & 4.3   & 49.9  & 4.0  \\
    \multirow{2}[0]{*}{\textbf{MaxRényi-K\% }} & \textbf{inst} & 51.5  & 9.3   & 54.8  & 11.0  & 52.9  & 6.3   & 51.7  & 7.3  \\
          & \textbf{desp} & 54.3  & 5.0   & 56.8  & 6.3   & 53.0  & 5.0   & 52.2  & 7.7  \\
    \multirow{2}[0]{*}{\textbf{ModRényi}} & \textbf{inst} & 52.4  & 7.3   & 53.1  & 7.7   & 48.8  & 8.0   & 54.8  & 7.3  \\
          & \textbf{desp} & 53.7  & 3.3   & 53.6  & 11.7  & 54.9  & 7.0   & 56.1  & 5.3  \\
\multicolumn{2}{c|}{\textbf{Image-only Inference}} & 52.5  & 4.6   & 51.7  & 8.3   & 53.3  & 7.1   & 50.5  & 6.1  \\

    \multicolumn{2}{c|}{\textbf{Blind Classifier (ours)}} & 49.0  & 8.3   & 49.0  & 8.3   & 49.0  & 8.3   & 49.0  & 8.3  \\
    \bottomrule[1.5pt]
    \end{tabular}%
    }
  \label{tab:coco}%
\end{table}%

Table~\ref{tab:coco} reports the performance of MI methods on COCO across four VLMs. AUCs hover around 50\% (random guessing), never exceeding 60\%, while TPR@5\%FPR remains below 10\% in most cases—indicating poor separability between members and non-members.
Table~\ref{tab:unbias} presents results on the instruction-tuning datasets of LLaVA OneVision and Molmo. Although these datasets are introduced at later training stages—where catastrophic forgetting is expected to be less severe—separability remains weak. 
\begin{table}[htbp]
  \centering
  \caption{MI Methods Performance on Debiased Model-Specific MI Datasets.}
  \resizebox{\linewidth}{!}{
    \begin{tabular}{cc|cccccc|cccccc}
    \toprule[1.5pt]
    \multicolumn{2}{c|}{\multirow{3}[0]{*}{\textbf{Method}}} & \multicolumn{6}{c|}{\textbf{LLaVA-OneVision}}         & \multicolumn{6}{c}{\textbf{Molmo}} \\
    \cmidrule(lr){3-8} \cmidrule(lr){9-14}
    \multicolumn{2}{c|}{} & \multicolumn{2}{c}{\textbf{ChartQA}} & \multicolumn{2}{c}{\textbf{DocVQA}} & \multicolumn{2}{c|}{\textbf{InfoVQA}} & \multicolumn{2}{c}{\textbf{PixChart}} & \multicolumn{2}{c}{\textbf{PixDigram}} & \multicolumn{2}{c}{\textbf{PixTable}} \\
    \cmidrule(lr){3-4} \cmidrule(lr){5-6} \cmidrule(lr){7-8} \cmidrule(lr){9-10} \cmidrule(lr){11-12} \cmidrule(lr){13-14}
\multicolumn{2}{c|}{} & \textbf{AUC} & \textbf{TPR} & \textbf{AUC} & \textbf{TPR} & \textbf{AUC} & \textbf{TPR} & \textbf{AUC} & \textbf{TPR} & \textbf{AUC} & \textbf{TPR} & \textbf{AUC} &\textbf{TPR} \\

    \midrule
    \multirow{2}[0]{*}{\textbf{Perpl}} & \textbf{in} & 47.8  & 2.0   & 47.8  & 1.0   & 39.3  & 1.4   & 48.8  & 1.7   & 49.3  & 3.0   & 51.0  & 6.0  \\
          & \textbf{de} & 53.9  & 4.3   & 54.2  & 2.7   & 50.7  & 4.8   & 53.1  & 5.0   & 52.2  & 9.3   & 46.7  & 5.0  \\
          \cmidrule{1-2}
    \multirow{2}[0]{*}{\textbf{\shortstack{Max\\Gap}}} & \textbf{in} & 41.7  & 2.7   & 57.9  & 8.7   & 56.4  & 10.2  & 49.7  & 6.7   & 52.1  & 4.7   & 51.9  & 8.3  \\
          & \textbf{de} & 55.1  & 9.0   & 55.9  & 6.3   & 50.4  & 4.8   & 50.5  & 4.7   & 49.2  & 3.7   & 47.4  & 4.0  \\
          \cmidrule{1-2}
    \multirow{2}[0]{*}{\textbf{\shortstack{Min\\K\%}}} & \textbf{in} & 52.9  & 5.3   & 52.6  & 4.0   & 41.4  & 1.4   & 48.8  & 4.0   & 50.8  & 4.0   & 52.9  & 10.0  \\
          & \textbf{de} & 54.1  & 4.7   & 54.2  & 4.3   & 52.3  & 6.8   & 53.2  & 5.7   & 53.3  & 10.3  & 52.1  & 6.7  \\
          \cmidrule{1-2}
    \multirow{2}[0]{*}{\textbf{\shortstack{Min\\K\%++}}} & \textbf{in} & 53.7  & 8.0   & 50.5  & 4.0   & 58.7  & 10.2  & 52.6  & 7.7   & 48.6  & 4.0   & 50.4  & 7.7  \\
          & \textbf{de} & 53.9  & 7.0   & 52.4  & 6.7   & 52.8  & 5.4   & 47.1  & 4.3   & 50.4  & 5.3   & 47.0  & 4.7  \\
          \cmidrule{1-2}
    \multirow{2}[0]{*}{\textbf{\shortstack{Max\\Rényi}}} & \textbf{in} & 48.6  & 5.3   & 56.6  & 14.7  & 58.4  & 8.8   & 51.9  & 7.7   & 55.1  & 7.0   & 49.2  & 6.7  \\
          & \textbf{de} & 53.3  & 4.7   & 54.0  & 4.7   & 53.4  & 8.2   & 52.3  & 8.7   & 54.7  & 7.3   & 51.5  & 6.7  \\
          \cmidrule{1-2}
    \multirow{2}[0]{*}{\textbf{\shortstack{Mod\\Rényi}}} & \textbf{in} & 53.1  & 5.7   & 46.1  & 4.0   & 45.1  & 5.4   & 51.4  & 7.0   & 49.6  & 6.3   & 50.0  & 5.0  \\
          & \textbf{de} & 53.7  & 3.3   & 54.4  & 4.7   & 50.8  & 4.8   & 54.8  & 4.3   & 52.0  & 10.0  & 47.5  & 5.7  \\
          \cmidrule{1-2}
    \multicolumn{2}{c|}{\textbf{Img Infer}} & 48.0  & 4.3   & 53.5  & 9.7   & 58.4  & 9.4   & 54.5  & 5.7   & 50.3  & 7.0   & 46.1  & 4.5  \\
    \cmidrule{1-2}
    \multicolumn{2}{c|}{\textbf{Blind Clf}} & 57.7  & 8.3   & 54.9  & 6.2   & 57.0  & 7.2   & 48.4  & 1.7   & 48.5  & 6.8   & 57.1  & 9.8  \\
    \bottomrule[1.5pt]
    \end{tabular}%
    }
  \label{tab:unbias}%
\end{table}%

In contrast, as shown in \S~\ref{sec:distri_shift}, when distribution shifts are present, a simple classifier trained on just 300 samples can achieve near-perfect separation (AUC~\(\approx 100\%\)). This stark contrast highlights how subtle the true membership signal is in VLM outputs compared to distributional artifacts. 
In this context, a natural question arises: is membership inference on VLMs truly feasible?

\subsection{Probing the Envelope of MI Performance}
\label{sec:probe}
To assess the feasibility of VLM membership inference, we consider an idealized setting where the auditor has full white-box access to the model’s internal embeddings—the source of potential memorization signals. This enables us to examine whether members and non-members are separable in the representation space, and to estimate the theoretical upper bound of this separability, characterized by Bayes optimality \cite{hastie2009elements, garber1988bounds}.

\subsubsection{Probing the VLM Embedding Space}
Given a target image \( I \in (0, 255)^{3 \times H \times W} \), we extract all hidden states from the vision encoder and language decoder during description generation. For each layer \( v_i \) in the vision encoder, we collect token-wise visual features \( \{\mathbf{h}_v^1, \mathbf{h}_v^2, \ldots\} \in \mathbb{R}^{d_v} \); for each layer \( l_i \) in the language decoder, we record hidden states generated during next-token prediction \( t_{k+1} = p(\theta; I, t_{\leq k}) \), denoted as \( \{\mathbf{h}_l^1, \mathbf{h}_l^2, \ldots\} \in \mathbb{R}^{d_l} \).
To evaluate the separability between members and non-members in these hidden states, we first apply average pooling to obtain fixed-size embeddings of dimension \( d_v \) and \( d_l \) for visual and language tokens, and then adopt standard probing methods \cite{liu2024probing}, training both linear classifiers and multi-layer perceptrons (MLPs) to distinguish member (class 1) from non-member (class 0) samples.

To mitigate potential information loss from global pooling, we further introduce an attention pooling classifier that adaptively aggregates the most informative tokens:

\begin{equation}
p(y_i = 1 \mid u, x_{\leq i}) = \sigma(\mathbf{w}^\top \bar{\mathbf{h}}_i)
\label{eq:attn_pred}
\end{equation}

where the aggregated representation \( \bar{\mathbf{h}}_i \) is computed as:

\begin{equation}
\bar{\mathbf{h}}_i = \sum_{j=1}^{i} \alpha_{i,j} \mathbf{h}_j, \quad 
\alpha_{i,j} = \frac{\exp(\mathbf{q}^\top \mathbf{h}_j)}{\sum_{k=1}^{i} \exp(\mathbf{q}^\top \mathbf{h}_k)}
\label{eq:attn_weights}
\end{equation}

Here, \( \mathbf{q} \in \mathbb{R}^{d} \) is a learnable query vector that attends to informative tokens. This enables the classifier to capture fine-grained membership signals that may be distributed across tokens.

\begin{table*}[htbp]
  \centering
  \caption{BER and Performance of Probing Methods using Visual and Language Tokens on the Debiased COCO Dataset.}
  \resizebox{\linewidth}{!}{
    \begin{tabular}{cc|ccccccccccc}
    \toprule[1.5pt]
    \multirow{2}[0]{*}{\textbf{Model}} & \multirow{2}[0]{*}{\textbf{Modal}} & \multicolumn{2}{c}{\textbf{BER}} & \multicolumn{3}{c}{\textbf{Linear}} & \multicolumn{3}{c}{\textbf{MLP}} & \multicolumn{3}{c}{\textbf{Attention}} \\
    \cmidrule(lr){3-4} \cmidrule(lr){5-7} \cmidrule(lr){8-10} \cmidrule(lr){11-13}
          &       & \textbf{Original} & \textbf{Calibrated} & \textbf{ACC} & \textbf{AUC} & \textbf{TPR} & \textbf{ACC} & \textbf{AUC} & \textbf{TPR} & \textbf{ACC} & \textbf{AUC} & \textbf{TPR} \\
    \midrule
    \multirow{2}[0]{*}{\textbf{LLaVA-1.5}} & \textbf{Vision} & 33.8  & 22.3       & 53.3±3.8 & 55.4±2.1 & 8.0±2.2 & 53.6±3.4 & 55.3±0.3 & 10.8±3.0 & 52.2±5.3 & 53.4±2.8 & 4.5±1.5 \\
          & \textbf{Language} & 26.3  & N/A       & 50.0±3.1 & 52.2±5.6 & 11.5±4.6 & 50.6±2.6 & 52.2±5.2 & 8.8±6.7 & 48.6±2.7 & 51.2±4.9 & 9.2±3.0 \\
    \multirow{2}[0]{*}{\textbf{Cambrian}} & \textbf{Vision} & 36.8  & 22.3       & 52.8±2.4 & 50.4±3.5 & 5.3±1.4 & 52.2±3.7 & 50.1±3.0 & 3.6±2.6 & 50.0±2.4 & 50.7±1.5 & 11.1±4.9 \\
          & \textbf{Language} & 23.3  & N/A       & 55.0±5.6 & 54.1±5.8 & 11.7±3.3 & 51.9±2.6 & 52.2±4.3 & 5.9±5.9 & 53.9±5.5 & 52.9±6.6 & 13.5±4.7 \\
    \multirow{2}[0]{*}{\textbf{LLaVA-OneVision}} & \textbf{Vision} & 21.2  & 22.3       & 50.8±4.2 & 58.8±6.3 & 10.3±4.3 & 52.5±3.4 & 56.4±5.2 & 15.5±6.1 & 55.8±4.1 & 60.2±5.7 & 8.6±3.5 \\
          & \textbf{Language} & 23.2  & N/A       & 63.3±5.7 & 63.4±4.8 & 8.6±2.4 & 60.8±4.7 & 63.8±5.3 & 5.9±3.2 & 62.5±4.1 & 65.0±4.3 & 6.9±3.7 \\
    \multirow{2}[0]{*}{\textbf{Molmo}} & \textbf{Vision} & 25.5  &  22.3      & 55.3±3.7 & 56.3±4.2 & 13.0±1.3 & 52.8±2.1 & 53.4±4.2 & 14.0±6.3 & 51.7±5.6 & 55.6±3.9 & 15.3±4.8 \\
          & \textbf{Language} & 21.7  &  N/A      & 50.3±3.5 & 52.1±4.1 & 10.0±1.5 & 51.1±1.4 & 52.0±3.3 & 8.1±8.0 & 50.3±2.7 & 53.6±3.5 & 12.3±3.5 \\
    \bottomrule[1.5pt]
    \end{tabular}%
    }
  \label{tab:coco_linear}%
\end{table*}%

\subsubsection{Estimating MI Performance Against Bayes Optimality}
While the probing methods assume full white-box access to the VLM—far beyond what is feasible with real-world APIs—they remain empirical in nature. One might conjecture that, as probing improves, member and non-member samples may eventually become fully separable.
To investigate this, we consider the theoretical upper bound of MI: the irreducible error in distinguishing members from non-members based on VLM hidden representations, quantified by the \textit{Bayes error rate} (BER) \cite{hastie2009elements, garber1988bounds}. Formally, BER is defined as the expected misclassification rate of the Bayes-optimal classifier under the task distribution \( D \):

\begin{equation}
\beta_D = \mathbb{E}_{(x,y) \sim D} \left[ 1 - \max_k p(y = k \mid x) \right]
\label{eq:ber_def1}
\end{equation}

Alternatively, it can be interpreted as the minimal error rate achievable over all measurable functions \( h \):

\begin{equation}
\beta_D = \min_{\text{measurable} \ h} \mathbb{E}_{(x,y) \sim D} \left[ \mathbb{I}(h(x) \neq y) \right]
\label{eq:ber_def2}
\end{equation}

where \( \mathbb{I} \) is the indicator function. Since VLM feature spaces do not follow simple, tractable distributions that permit analytical computation of BER, we adopt an efficient approximation \cite{chen2023evaluating} to estimate BER in distinguishing membership in VLM hidden states.

Specifically, we compute pairwise \( \ell_2 \) distances between token features to construct an adjacency matrix \( \mathbf{A} \), where \( A_{ij} = 1 \) if \( \mathbf{x}_i \) and \( \mathbf{x}_j \) share the same label, and 0 otherwise. Connected components representing confident regions are identified via breadth-first search. Remaining unconnected samples are treated as uncertain and labeled using Label Spreading \cite{zhu2003semi}. BER is then estimated as the fraction of incorrect predictions among these uncertain samples:

\begin{equation}
\text{BER} = \frac{1}{n} \sum_{i \in \mathcal{U}} \mathbb{I}(\hat{y}_i \neq y_i)
\label{eq:ber_est}
\end{equation}

where \( \mathcal{U} \) denotes the set of uncertain samples, \( \hat{y}_i \) the predicted label, and \( y_i \) the ground-truth label.
Note that BER provides a \textit{highly optimistic} estimate of the lowest achievable error. The corresponding Bayes-optimal classifier is not accessible in practice, and BER does not account for generalization—it may reflect separability based on spurious, non-transferable features. 

\subsubsection{Experimental Results}

\begin{table}[htbp]
  \centering
  \caption{Performance of Probing Methods on Model-Specific Debiased Datasets.}
  \resizebox{\linewidth}{!}{
    \begin{tabular}{cc|cccccc}
    \toprule[1.5pt]
    \multirow{2}[0]{*}{\textbf{Dataset}} & \multirow{2}[0]{*}{\textbf{Method}} & \multicolumn{3}{c}{\textbf{Vision}} & \multicolumn{3}{c}{\textbf{Language}} \\
    \cmidrule(lr){3-5} \cmidrule(lr){6-8}
          &       & \textbf{ACC} & \textbf{AUC} & \textbf{TPR} & \textbf{ACC} & \textbf{AUC} & \textbf{TPR} \\
    \midrule
    \multirow{3}[0]{*}{\textbf{\shortstack{Chart\\QA}}} & \textbf{Linear} & 51.7±1.4 & 53.2±1.4 & 0.6±0.8 & 56.1±0.8 & 56.0±0.8 & 5.4±3.9 \\
          & \textbf{MLP} & 49.2±1.8 & 53.0±0.7 & 6.4±2.8 & 57.2±4.2 & 56.0±2.7 & 2.3±3.3 \\
          & \textbf{Attention} & 51.9±1.6 & 52.9±1.1 & 4.1±3.0 & 53.3±1.2 & 53.5±1.2 & 5.3±2.4 \\
    \cmidrule(lr){1-2}
    \multirow{3}[0]{*}{\textbf{\shortstack{Doc\\VQA}}} & \textbf{Linear} & 52.8±0.8 & 54.7±3.0 & 6.5±0.9 & 46.1±1.7 & 46.8±1.3 & 6.4±2.8 \\
          & \textbf{MLP} & 54.4±5.5 & 56.0±7.0 & 7.0±2.7 & 46.9±2.2 & 46.3±1.7 & 0.0±0.0 \\
          & \textbf{Attention} & 53.3±0.0 & 55.8±2.6 & 4.7±2.2 & 46.7±0.7 & 46.7±0.1 & 7.1±1.6 \\
    \cmidrule(lr){1-2}
    \multirow{3}[0]{*}{\textbf{\shortstack{Info\\VQA}}} & \textbf{Linear} & 66.4±3.8 & 73.3±3.3 & 25.5±7.8 & 67.5±0.8 & 71.3±3.6 & 15.3±3.8 \\
          & \textbf{MLP} & 64.4±4.0 & 69.4±0.5 & 22.0±8.5 & 68.1±3.4 & 73.0±4.1 & 14.5±10.7 \\
          & \textbf{Attention} & 69.5±3.5 & 74.2±3.6 & 29.7±4.4 & 66.1±1.4 & 72.0±2.3 & 9.6±2.4 \\
    \midrule
    \multirow{3}[0]{*}{\textbf{\shortstack{Pix\\Chart}}} & \textbf{Linear} & 53.3±2.5 & 50.8±4.2 & 5.8±4.4 & 55.3±3.9 & 56.8±4.5 & 4.2±2.4 \\
          & \textbf{MLP} & 51.9±5.2 & 49.8±5.3 & 7.1±1.3 & 56.7±4.8 & 57.4±4.1 & 2.5±3.5 \\
          & \textbf{Attention} & 54.7±1.0 & 50.3±0.5 & 2.3±1.6 & 54.4±3.1 & 55.8±3.7 & 3.0±3.2 \\
    \cmidrule(lr){1-2}
    \multirow{3}[0]{*}{\textbf{\shortstack{Pix\\Digram}}} & \textbf{Linear} & 50.8±5.1 & 48.7±4.2 & 6.5±2.3 & 53.9±6.1 & 50.3±6.1 & 10.1±1.1 \\
          & \textbf{MLP} & 46.9±3.1 & 47.1±4.4 & 4.2±1.7 & 48.9±4.4 & 48.9±4.4 & 4.9±4.6 \\
          & \textbf{Attention} & 51.9±3.1 & 54.6±1.4 & 10.2±6.2 & 48.6±2.7 & 49.9±3.0 & 8.3±2.3 \\
    \cmidrule(lr){1-2}
    \multirow{3}[0]{*}{\textbf{\shortstack{Pix\\Table}}} & \textbf{Linear} & 46.4±4.0 & 49.2±5.0 & 5.3±1.3 & 46.4±1.6 & 46.1±0.9 & 5.9±4.6 \\
          & \textbf{MLP} & 51.1±4.4 & 49.3±4.5 & 5.3±1.4 & 46.7±2.5 & 46.5±2.6 & 0.6±0.9 \\
          & \textbf{Attention} & 50.0±6.2 & 50.0±4.9 & 3.5±3.7 & 45.3±1.4 & 45.7±1.6 & 8.2±4.5 \\
    \bottomrule[1.5pt]
    \end{tabular}%
    }
  \label{tab:unbias-linear}%
\end{table}%
Tables~\ref{tab:coco_linear} and \ref{tab:unbias-linear} report the performance of three probing methods applied to visual and language tokens across our unbiased datasets. In most cases, both accuracy and AUC remain below 65\%, and the sophisticated attention pooling classifier fails to yield noticeable improvements, indicating that even at the source of memorization signals—the internal representations of VLMs—members and non-members remain weakly separable.
Furthermore, in most cases, BER falls between 20\% and 30\%, implying that the theoretical upper bound for MI is only around 70\%. Notably, BER is an optimistic estimate that does not reflect practical generalization.
As a sanity check, we project the same images into the feature space of an ImageNet-pretrained EfficientNet. Although all samples are non-members from EfficientNet’s perspective, the calibrated BER (Cal) in Table~\ref{tab:coco_linear} and \ref{tab:BER} still ranges from 20\% to 30\%. This suggests that even with genuine membership signals, VLM representations offer only marginally better separation.

\begin{table}[htbp]
  \centering
  \caption{BER on Debiased Model-Specific MI Datasets.}
  \resizebox{\linewidth}{!}{
    \begin{tabular}{c|cccc||c|cccc}
    \toprule[1.5pt]
    \multirow{2}[0]{*}{\textbf{Dataset}} & \multicolumn{2}{c}{\textbf{Vision}} & \multicolumn{2}{c||}{\textbf{Language}} & \multirow{2}[0]{*}{\textbf{Dataset}} & \multicolumn{2}{c}{\textbf{Vision}} & \multicolumn{2}{c}{\textbf{Language}} \\
          & \textbf{Ori} & \textbf{Cal} & \textbf{Ori} & \textbf{Cal} &       & \textbf{Ori} & \textbf{Cal} & \textbf{Ori} & \textbf{Cal} \\
    \midrule
    \textbf{ChartQA} & 24.0  & 32.7  & 23.0  & N/A  & \textbf{PixChart} & 32.5  & 28.3  & 26.2  & N/A  \\
    \textbf{DocVQA} & 17.5  & 25.2  & 20.3  & N/A & \textbf{PixDigram} & 33.3  & 27.5  & 26.2  & N/A  \\
    \textbf{InfoVQA} & 20.5  & 20.9  & 15.9  & N/A  & \textbf{PixTable} & 30.7  & 27.0  & 22.2  & N/A  \\
    \bottomrule[1.5pt]
    \end{tabular}%
    }
  \label{tab:BER}%
\end{table}%
\subsubsection{Ablations}
Our default setting uses final-layer features of 7B-scale VLMs. To examine whether these choices limit separability, we conduct ablations on COCO with the LLaVA OneVision family, varying three key factors: layer depth, model scale, and output length.
\begin{figure}[hbtp] 
    \centering
    \includegraphics[width=\linewidth]{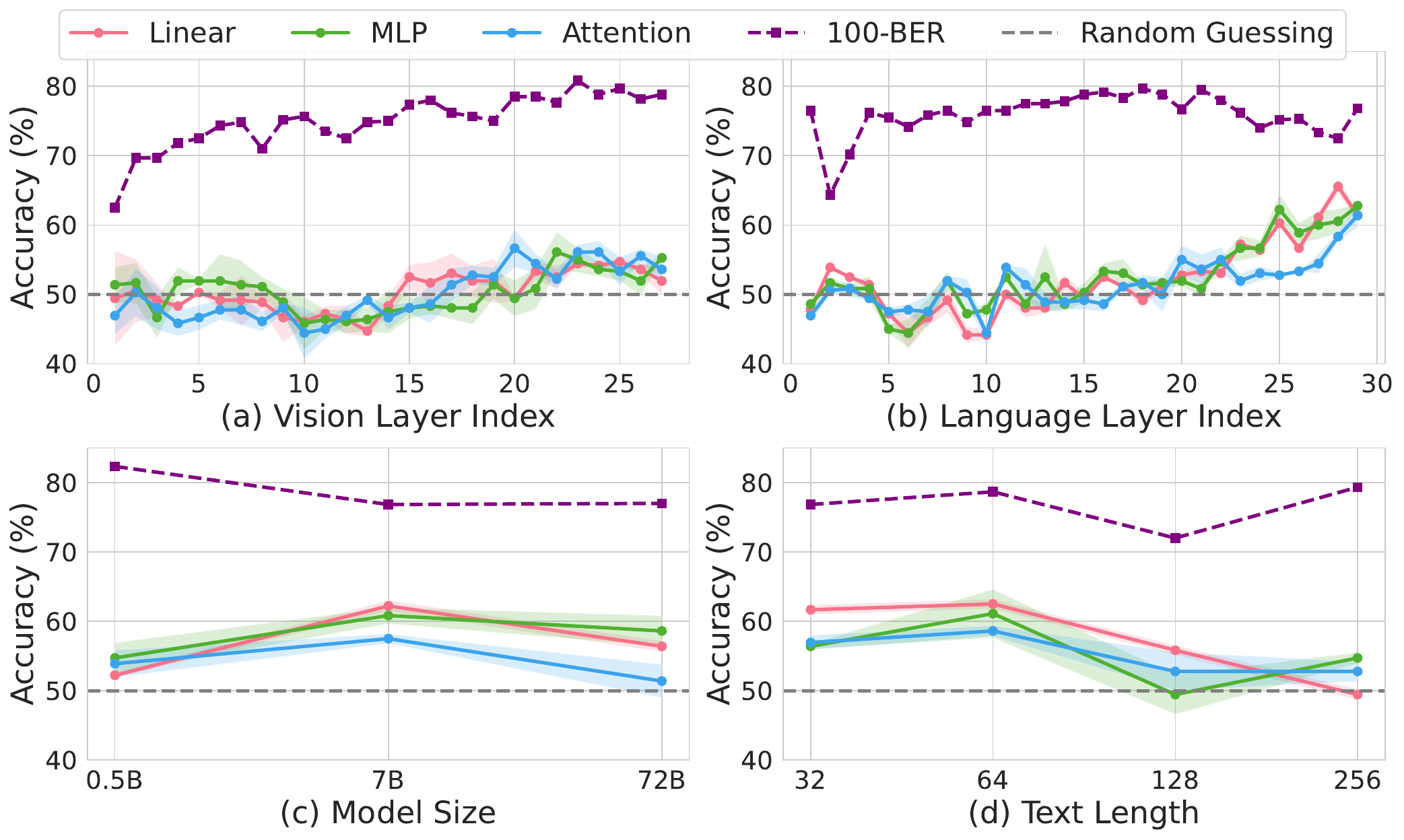} 
    \caption{Ablation Performance of Probing Methods(LLaVA-ov on COCO).} 
    \label{fig:combined_plots} 
\end{figure}
Figures~\ref{fig:combined_plots}(a)(b) show the accuracy of pooling classifiers and the corresponding Bayes optimality (\(100 - \text{BER}\)) across different layers. While deeper layers offer slightly improved separability, results remain near random guessing. Figure~\ref{fig:combined_plots}(c) evaluates models ranging from 0.5B to 72B parameters; despite the substantial increase in capacity, the largest model still fails to distinguish membership. Figure~\ref{fig:combined_plots}(d) explores the effect of output length, revealing that generating longer descriptions also does not improve separability.

\begin{tcolorbox}[colback=gray!5!white,
                  colframe=teal!80!black,
                  coltitle=black,
                  fonttitle=\bfseries\itshape,
                  enhanced,
                  sharp corners=southwest,
                  drop shadow,
                  after=\vspace{0pt}]
\textbf{Finding 3:} Even with oracle access to internal states, membership signals remain subtle. Probing classifiers on hidden features yield only marginal gains, and Bayes optimality remains low, indicating minimal room for improvement.
\end{tcolorbox}

\section{When Does VLM MI Become Feasible?}
\label{sec:when}
We analyze why MI struggles in large VLMs and construct targeted scenarios to mitigate these challenges. Surprisingly, we find MI becomes feasible in these realistic auditing settings.
\subsection{Finetuning on Downstream Tasks}
\subsubsection{Challenge: Minimal Overfitting}
Traditional MI typically targets models trained for many epochs on specific downstream tasks. In contrast, LLMs and VLMs adopt general-purpose training objectives and often see each example only once~\cite{muennighoff2023scaling}. Early VLMs like LLaVA-1 \cite{liu2023visual} performed multi-epoch training (e.g., 3), but recent models have shifted toward data scaling—training on massive, high-quality datasets for a single epoch—which enables generalization without severe overfitting~\cite{tirumala2022memorization, antoniades2025generalization}. Moreover, due to the sheer data volume, early examples are frequently forgotten during training~\cite{jagielskimeasuring}. Theoretically, as dataset size grows, model behavior converges across seen and unseen samples~\cite{mainidataset}.
In \S~\ref{sec:probe}, we observe unexpectedly better membership separability in LLaVA OneVision on COCO, despite COCO being used during the more forgettable alignment phase. Upon closer inspection, we identify substantial image overlap among MS COCO~\cite{lin2014microsoft}, COCO Caption~\cite{chen2015microsoft}, and RefCOCO~\cite{yu2016modeling} in the training corpus. We hypothesize that this duplication amplifies MI signals and ask: can stronger MI emerge when VLMs are fine-tuned for multiple epochs?

\subsubsection{Scenario: Finetuning on Downstream Tasks}
To ensure i.i.d. splits, we continue using instruction-tuning images with random partitions. Since test splits provide only short text answers, we generate image descriptions using LLaVA OneVision-7B, then fine-tune LLaVA-1.5-7B with LoRA \cite{hu2022lora} for 10 epochs.
\begin{table}[htbp]
  \centering
  \caption{MI Performance on LoRA LLava-1.5 (gt Refers to Ground Truth).}
  \resizebox{\linewidth}{!}{
    \begin{tabular}{c|cccccccc}
    \toprule[1.5pt]
    \multirow{3}[0]{*}{\textbf{Method}} & \multicolumn{4}{c}{\textbf{ChartQA}} & \multicolumn{4}{c}{\textbf{DocVQA}} \\
    \cmidrule(lr){2-5}\cmidrule(lr){6-9}
          & \multicolumn{2}{c}{\textbf{epoch 10 w/o gt}} & \multicolumn{2}{c}{\textbf{epoch 3 w/ gt}} & \multicolumn{2}{c}{\textbf{epoch10 w/o gt}} & \multicolumn{2}{c}{\textbf{epoch 3 w/ gt}} \\
    \cmidrule(lr){2-3}\cmidrule(lr){4-5}\cmidrule(lr){6-7}\cmidrule(lr){8-9}
          & \textbf{AUC} & \textbf{TPR} & \textbf{AUC} & \textbf{TPR} & \textbf{AUC} & \textbf{TPR} & \textbf{AUC} & \textbf{TPR} \\
    \midrule
    \textbf{Perplexity} & 60.8  & 16.3  & 78.2  & 8.3   & 65.2  & 18.7  & 75.3  & 19.0  \\
    \textbf{Min-K\%} & 60.9  & 16.7  & 78.7  & 14.3  & 65.2  & 18.7  & 76.2  & 26.7  \\
    \textbf{ModRényi} & 60.8  & 16.7  & 79.3  & 14.3  & 64.9  & 18.0  & 76.7  & 20.7  \\
    \bottomrule[1.5pt]
    \end{tabular}%
    }
  \label{tab:lora}%
\end{table}%
As shown in Table~\ref{tab:lora}, MI performance improves markedly by the 10th epoch, with AUCs surpassing 0.6. Notably, traditional MI approaches such as perplexity emerge as competitive baselines, indicating substantial room for improvement. This setting reflects realistic auditing scenarios, where privacy-sensitive or proprietary data (e.g., in medical VQA \cite{li2023llava}) are commonly involved during fine-tuning.
\subsection{Access to Ground-Truth Text}
\subsubsection{Challenge: Lack of Ground-Truth Captions}
In LLM-based MI, auditors typically query the model with a suspicious text $s = (\text{token}_1, \text{token}_2, \ldots, \text{token}_t)$ and records the output probabilities $p(\text{token}_k \mid \theta, \text{token}_{<k})$ via teacher forcing for inference.
However, extending this to VLMs poses a key challenge: the training caption for a suspicious image is usually inaccessible. Web-scraped image-text pairs are noisy and short, while modern VLMs are trained on high-quality, relabeled captions \cite{li2025llava, deitke2025molmo}. As a result, in most proprietary black-box VLMs, auditors cannot access the original training text.
To bypass this, MI methods like VL-MIA \cite{li2024membership} rely on the VLM’s own generated captions. However, these are merely \textit{pseudo ground-truths}. Due to snowballing prediction errors in autoregressive decoding \cite{bachmann2024pitfalls}, the outputs increasingly diverge from the true training text. Given this gap, an open question is: would MI attacks be more effective if ground-truth captions were available?
\subsubsection{Scenario: Access to Ground-Truth Text}
Due to the lack of i.i.d. train/test text splits in open datasets, we continue using the fine-tuned LLaVA-1.5-7B and evaluate the 3rd-epoch checkpoint with ground-truth text in Table~\ref{tab:lora}. Since LoRA updates less than 3\% of parameters, free-form generation at early epochs yields near-random MI results. In contrast, ground-truth text enables significantly stronger MI, with AUCs nearing 0.8—outperforming free-form results even after 10 epochs. This underscores MI’s value in detecting test set contamination \cite{oren2023proving} and verifying model ownership via curated samples \cite{zhu2024reliable}.
\subsection{Aggregation-Based Set Inference}
\subsubsection{Challenge: Diverse Intrinsic Image Attributes}
\S~\ref{sec:shortcut} shows that MI is sensitive to distribution shifts. The diverse sources of VLM training data introduce natural image-level variation—quality, noise, object count, and complexity—that can dominate token distributions, overshadowing subtle membership signals. However, when member and non-member images are drawn i.i.d., such high-variance factors may average out at the set level. This motivates the question: can aggregated MI signals across multiple images enable reliable set-level inference?

\subsubsection{Scenario: Aggregation-Based Set Inference}
\begin{figure}[hbtp] 
    \centering
    \includegraphics[width=\linewidth]{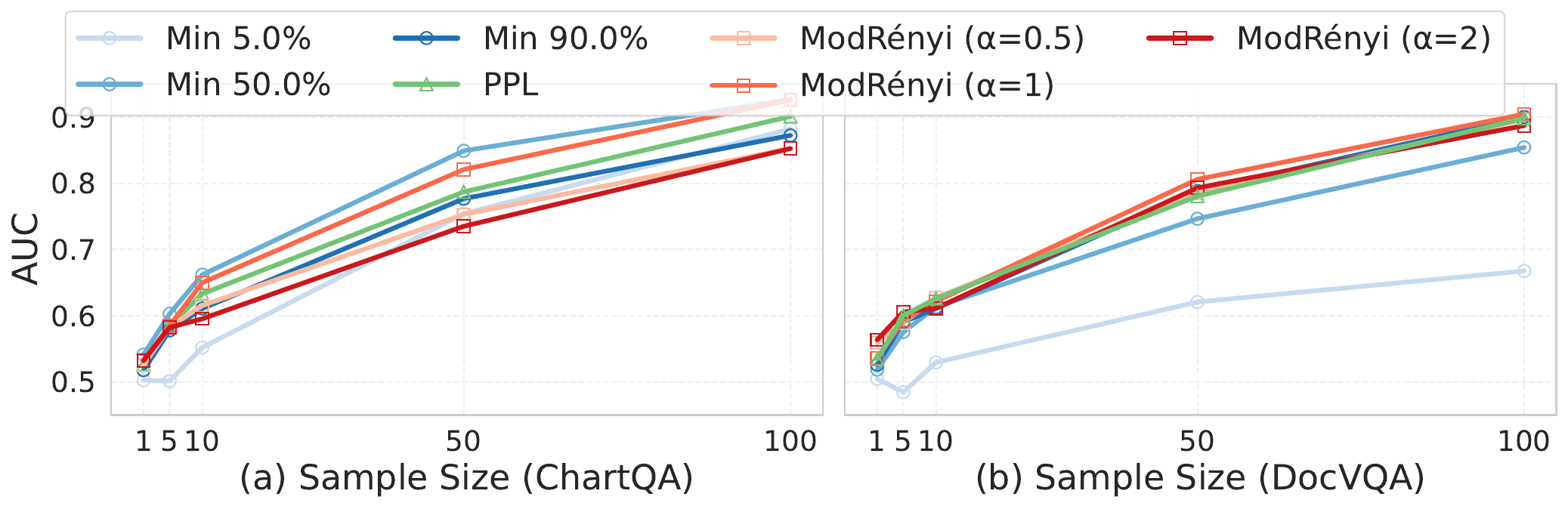} 
    \caption{MI Performance in Aggregation-based Set Inference on LLaVA-ov.} 
    \label{fig:line2} 
\end{figure}
We treat a group of images as a single MI unit by averaging MI scores. Inspired by bootstrapping \cite{efron2021computer}, we sample 1,000 sets with replacement from member and non-member pools, varying set sizes from 1 to 100. Figure~\ref{fig:line2} shows MI AUCs on LLaVA-OneVision. Even when single-image AUCs are only slightly above chance, set-level performance improves markedly. This approach is well-suited for auditing image collections—e.g., social media albums or artist portfolios \cite{ha2024organic}—and detecting unauthorized use of proprietary datasets \cite{du2024sok}. Notably, small per-image gains can translate into substantial set-level improvements, motivating further development of MI techniques.

\begin{tcolorbox}[colback=gray!5!white,
                  colframe=teal!80!black,
                  coltitle=black,
                  fonttitle=\bfseries\itshape,
                  enhanced,
                  sharp corners=southwest,
                  drop shadow,
                  after=\vspace{0pt}]
\textbf{Finding 4:} VLM MI becomes feasible when: (1) fine-tuning induces overfitting; (2) ground-truth text is available; or (3) predictions are aggregated—key scenarios for auditing test set contamination and collection copyright infringement.
\end{tcolorbox}

\section{Conclusion}
In this work, we identify a critical issue in current MI benchmarks for large VLMs: distribution shifts between member and non-member images introduce spurious shortcuts that overshadow true membership signals. We analyze these shifts and propose a principled metric to quantify them, enabling practical MI auditing.
To build an unbiased testbed, we reconstruct i.i.d. member/non-member splits from open-source VLMs. Under this setting, existing MI methods perform only slightly above chance. We further assess the theoretical upper bound of membership separability and find a high irreducible Bayes error, underscoring the fundamental difficulty of MI on VLMs.
Despite these challenges, we identify three practical scenarios where MI remains feasible and valuable for auditing: fine-tuning, access to ground-truth text, and aggregation across samples.
Future work will explore additional viable settings, design stronger MI methods, and extend our study to closed-source VLMs, such as auditing fine-tuned GPT-4o via API access.


\newpage
\bibliographystyle{ACM-Reference-Format}
\balance
\bibliography{sample-base}


\begin{thebibliography}{67}


\ifx \showCODEN    \undefined \def \showCODEN     #1{\unskip}     \fi
\ifx \showISBNx    \undefined \def \showISBNx     #1{\unskip}     \fi
\ifx \showISBNxiii \undefined \def \showISBNxiii  #1{\unskip}     \fi
\ifx \showISSN     \undefined \def \showISSN      #1{\unskip}     \fi
\ifx \showLCCN     \undefined \def \showLCCN      #1{\unskip}     \fi
\ifx \shownote     \undefined \def \shownote      #1{#1}          \fi
\ifx \showarticletitle \undefined \def \showarticletitle #1{#1}   \fi
\ifx \showURL      \undefined \def \showURL       {\relax}        \fi
\providecommand\bibfield[2]{#2}
\providecommand\bibinfo[2]{#2}
\providecommand\natexlab[1]{#1}
\providecommand\showeprint[2][]{arXiv:#2}

\bibitem[Anthropic(2024)]%
        {anthropic2024claude35sonnet}
\bibfield{author}{\bibinfo{person}{Anthropic}.} \bibinfo{year}{2024}\natexlab{}.
\newblock \bibinfo{title}{Introducing Claude 3.5 Sonnet}.
\newblock
\urldef\tempurl%
\url{https://www.anthropic.com/news/claude-3-5-sonnet}
\showURL{%
\tempurl}


\bibitem[Antoniades et~al\mbox{.}(2025)]%
        {antoniades2025generalization}
\bibfield{author}{\bibinfo{person}{Antonis Antoniades}, \bibinfo{person}{Xinyi Wang}, \bibinfo{person}{Yanai Elazar}, \bibinfo{person}{Alfonso Amayuelas}, \bibinfo{person}{Alon Albalak}, \bibinfo{person}{Kexun Zhang}, {and} \bibinfo{person}{William~Yang Wang}.} \bibinfo{year}{2025}\natexlab{}.
\newblock \showarticletitle{Generalization v.s. Memorization: Tracing Language Models’ Capabilities Back to Pretraining Data}. In \bibinfo{booktitle}{\emph{Proceedings of the 12th International Conference on Learning Representations (ICLR)}}.
\newblock


\bibitem[Bachmann and Nagarajan(2024)]%
        {bachmann2024pitfalls}
\bibfield{author}{\bibinfo{person}{Gregor Bachmann} {and} \bibinfo{person}{Vaishnavh Nagarajan}.} \bibinfo{year}{2024}\natexlab{}.
\newblock \showarticletitle{The Pitfalls of Next-Token Prediction}. In \bibinfo{booktitle}{\emph{International Conference on Machine Learning}}. PMLR, \bibinfo{pages}{2296--2318}.
\newblock


\bibitem[Bai et~al\mbox{.}(2025)]%
        {bai2025qwen2}
\bibfield{author}{\bibinfo{person}{Shuai Bai}, \bibinfo{person}{Keqin Chen}, \bibinfo{person}{Xuejing Liu}, \bibinfo{person}{Jialin Wang}, \bibinfo{person}{Wenbin Ge}, \bibinfo{person}{Sibo Song}, \bibinfo{person}{Kai Dang}, \bibinfo{person}{Peng Wang}, \bibinfo{person}{Shijie Wang}, \bibinfo{person}{Jun Tang}, {et~al\mbox{.}}} \bibinfo{year}{2025}\natexlab{}.
\newblock \showarticletitle{Qwen2. 5-vl technical report}.
\newblock \bibinfo{journal}{\emph{arXiv preprint arXiv:2502.13923}} (\bibinfo{year}{2025}).
\newblock


\bibitem[Bischoff et~al\mbox{.}(2024)]%
        {bischoffpractical}
\bibfield{author}{\bibinfo{person}{Sebastian Bischoff}, \bibinfo{person}{Alana Darcher}, \bibinfo{person}{Michael Deistler}, \bibinfo{person}{Richard Gao}, \bibinfo{person}{Franziska Gerken}, \bibinfo{person}{Manuel Gloeckler}, \bibinfo{person}{Lisa Haxel}, \bibinfo{person}{Jaivardhan Kapoor}, \bibinfo{person}{Janne~K Lappalainen}, \bibinfo{person}{Jakob~H Macke}, {et~al\mbox{.}}} \bibinfo{year}{2024}\natexlab{}.
\newblock \showarticletitle{A Practical Guide to Sample-based Statistical Distances for Evaluating Generative Models in Science}.
\newblock \bibinfo{journal}{\emph{Transactions on Machine Learning Research}} (\bibinfo{year}{2024}).
\newblock


\bibitem[Breiman(2001)]%
        {breiman2001random}
\bibfield{author}{\bibinfo{person}{Leo Breiman}.} \bibinfo{year}{2001}\natexlab{}.
\newblock \showarticletitle{Random forests}.
\newblock \bibinfo{journal}{\emph{Machine learning}}  \bibinfo{volume}{45} (\bibinfo{year}{2001}), \bibinfo{pages}{5--32}.
\newblock


\bibitem[Carlini et~al\mbox{.}(2021)]%
        {carlini2021extracting}
\bibfield{author}{\bibinfo{person}{Nicholas Carlini}, \bibinfo{person}{Florian Tramer}, \bibinfo{person}{Eric Wallace}, \bibinfo{person}{Matthew Jagielski}, \bibinfo{person}{Ariel Herbert-Voss}, \bibinfo{person}{Katherine Lee}, \bibinfo{person}{Adam Roberts}, \bibinfo{person}{Tom Brown}, \bibinfo{person}{Dawn Song}, \bibinfo{person}{Ulfar Erlingsson}, {et~al\mbox{.}}} \bibinfo{year}{2021}\natexlab{}.
\newblock \showarticletitle{Extracting training data from large language models}. In \bibinfo{booktitle}{\emph{30th USENIX security symposium (USENIX Security 21)}}. \bibinfo{pages}{2633--2650}.
\newblock


\bibitem[Chen et~al\mbox{.}(2024)]%
        {chen2024sharegpt4v}
\bibfield{author}{\bibinfo{person}{Lin Chen}, \bibinfo{person}{Jinsong Li}, \bibinfo{person}{Xiaoyi Dong}, \bibinfo{person}{Pan Zhang}, \bibinfo{person}{Conghui He}, \bibinfo{person}{Jiaqi Wang}, \bibinfo{person}{Feng Zhao}, {and} \bibinfo{person}{Dahua Lin}.} \bibinfo{year}{2024}\natexlab{}.
\newblock \showarticletitle{Sharegpt4v: Improving large multi-modal models with better captions}. In \bibinfo{booktitle}{\emph{European Conference on Computer Vision}}. Springer, \bibinfo{pages}{370--387}.
\newblock


\bibitem[Chen et~al\mbox{.}(2023)]%
        {chen2023evaluating}
\bibfield{author}{\bibinfo{person}{Qingqiang Chen}, \bibinfo{person}{Fuyuan Cao}, \bibinfo{person}{Ying Xing}, {and} \bibinfo{person}{Jiye Liang}.} \bibinfo{year}{2023}\natexlab{}.
\newblock \showarticletitle{Evaluating classification model against Bayes error rate}.
\newblock \bibinfo{journal}{\emph{IEEE Transactions on Pattern Analysis and Machine Intelligence}} \bibinfo{volume}{45}, \bibinfo{number}{8} (\bibinfo{year}{2023}), \bibinfo{pages}{9639--9653}.
\newblock


\bibitem[Chen et~al\mbox{.}(2015)]%
        {chen2015microsoft}
\bibfield{author}{\bibinfo{person}{Xinlei Chen}, \bibinfo{person}{Hao Fang}, \bibinfo{person}{Tsung-Yi Lin}, \bibinfo{person}{Ramakrishna Vedantam}, \bibinfo{person}{Saurabh Gupta}, \bibinfo{person}{Piotr Doll{\'a}r}, {and} \bibinfo{person}{C~Lawrence Zitnick}.} \bibinfo{year}{2015}\natexlab{}.
\newblock \showarticletitle{Microsoft coco captions: Data collection and evaluation server}.
\newblock \bibinfo{journal}{\emph{arXiv preprint arXiv:1504.00325}} (\bibinfo{year}{2015}).
\newblock


\bibitem[Das et~al\mbox{.}(2024)]%
        {das2024blind}
\bibfield{author}{\bibinfo{person}{Debeshee Das}, \bibinfo{person}{Jie Zhang}, {and} \bibinfo{person}{Florian Tram{\`e}r}.} \bibinfo{year}{2024}\natexlab{}.
\newblock \showarticletitle{Blind baselines beat membership inference attacks for foundation models}.
\newblock \bibinfo{journal}{\emph{arXiv preprint arXiv:2406.16201}} (\bibinfo{year}{2024}).
\newblock


\bibitem[Deitke et~al\mbox{.}(2025)]%
        {deitke2025molmo}
\bibfield{author}{\bibinfo{person}{Matt Deitke}, \bibinfo{person}{Christopher Clark}, \bibinfo{person}{Sangho Lee}, {and} \bibinfo{person}{et al.}} \bibinfo{year}{2025}\natexlab{}.
\newblock \showarticletitle{Molmo and PixMo: Open Weights and Open Data for State-of-the-Art Vision-Language Models}. In \bibinfo{booktitle}{\emph{Proceedings of the IEEE/CVF Conference on Computer Vision and Pattern Recognition (CVPR)}}.
\newblock


\bibitem[Du et~al\mbox{.}(2024)]%
        {du2024sok}
\bibfield{author}{\bibinfo{person}{Linkang Du}, \bibinfo{person}{Xuanru Zhou}, \bibinfo{person}{Min Chen}, \bibinfo{person}{Chusong Zhang}, \bibinfo{person}{Zhou Su}, \bibinfo{person}{Peng Cheng}, \bibinfo{person}{Jiming Chen}, {and} \bibinfo{person}{Zhikun Zhang}.} \bibinfo{year}{2024}\natexlab{}.
\newblock \showarticletitle{SoK: Dataset Copyright Auditing in Machine Learning Systems}. In \bibinfo{booktitle}{\emph{2025 IEEE Symposium on Security and Privacy (SP)}}. IEEE Computer Society, \bibinfo{pages}{25--25}.
\newblock


\bibitem[Duarte et~al\mbox{.}(2024)]%
        {duarte2024cop}
\bibfield{author}{\bibinfo{person}{Andr{\'e}~V Duarte}, \bibinfo{person}{Xuandong Zhao}, \bibinfo{person}{Arlindo~L Oliveira}, {and} \bibinfo{person}{Lei Li}.} \bibinfo{year}{2024}\natexlab{}.
\newblock \showarticletitle{DE-COP: detecting copyrighted content in language models training data}. In \bibinfo{booktitle}{\emph{Proceedings of the 41st International Conference on Machine Learning}}. \bibinfo{pages}{11940--11956}.
\newblock


\bibitem[Efron and Hastie(2021)]%
        {efron2021computer}
\bibfield{author}{\bibinfo{person}{Bradley Efron} {and} \bibinfo{person}{Trevor Hastie}.} \bibinfo{year}{2021}\natexlab{}.
\newblock \bibinfo{booktitle}{\emph{Computer age statistical inference, student edition: algorithms, evidence, and data science}}. Vol.~\bibinfo{volume}{6}.
\newblock \bibinfo{publisher}{Cambridge University Press}.
\newblock


\bibitem[Engstrom et~al\mbox{.}(2020)]%
        {engstrom2020identifying}
\bibfield{author}{\bibinfo{person}{Logan Engstrom}, \bibinfo{person}{Andrew Ilyas}, \bibinfo{person}{Shibani Santurkar}, \bibinfo{person}{Dimitris Tsipras}, \bibinfo{person}{Jacob Steinhardt}, {and} \bibinfo{person}{Aleksander Madry}.} \bibinfo{year}{2020}\natexlab{}.
\newblock \showarticletitle{Identifying statistical bias in dataset replication}. In \bibinfo{booktitle}{\emph{International Conference on Machine Learning}}. PMLR, \bibinfo{pages}{2922--2932}.
\newblock


\bibitem[{Flickr}(2025)]%
        {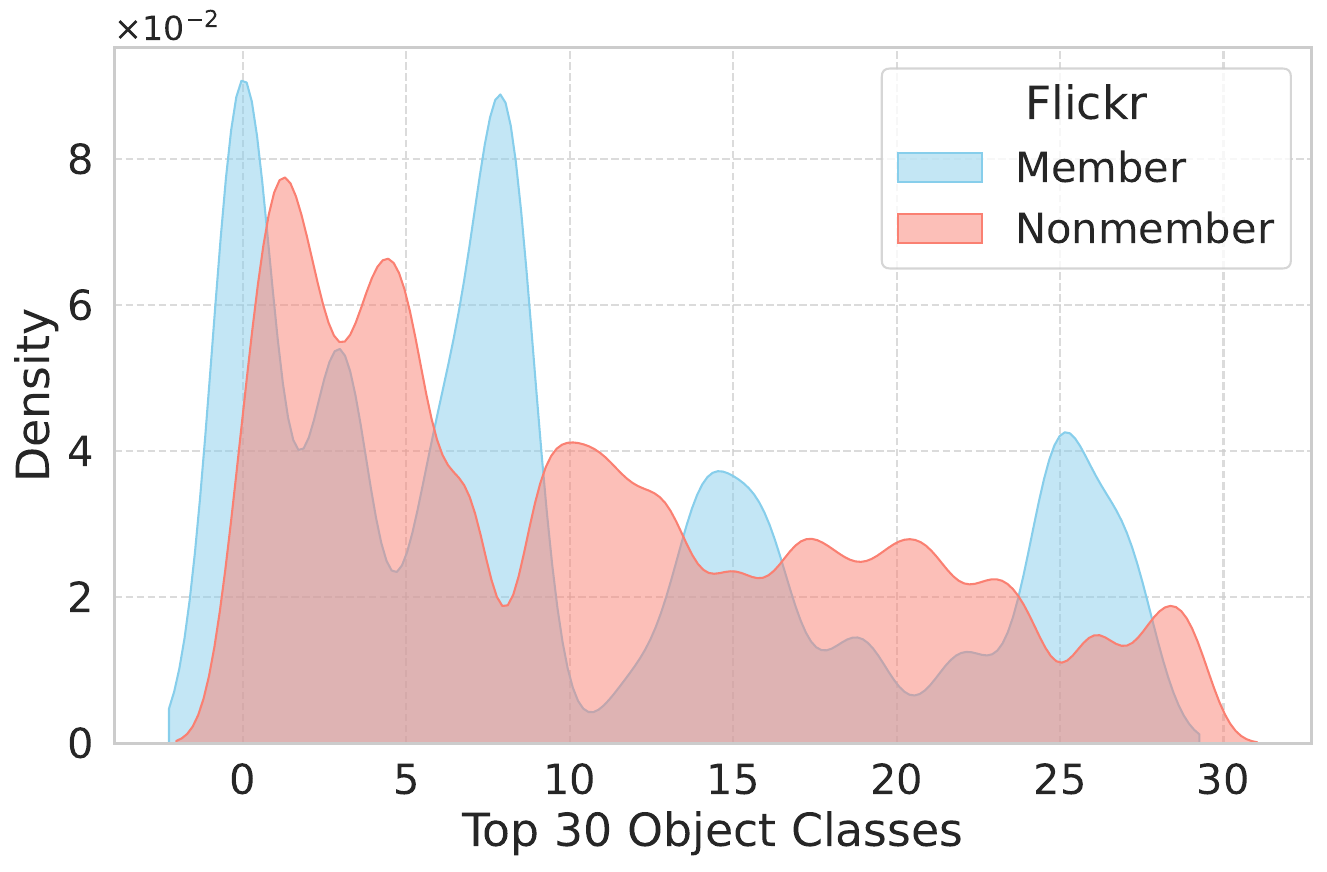}
\bibfield{author}{\bibinfo{person}{{Flickr}}.} \bibinfo{year}{2025}\natexlab{}.
\newblock \bibinfo{title}{Flickr: Online photo management and sharing application}.
\newblock \bibinfo{howpublished}{\url{https://www.flickr.com}}.
\newblock


\bibitem[Gadre et~al\mbox{.}(2023)]%
        {gadre2023datacomp}
\bibfield{author}{\bibinfo{person}{Samir~Yitzhak Gadre}, \bibinfo{person}{Gabriel Ilharco}, \bibinfo{person}{Alex Fang}, \bibinfo{person}{Jonathan Hayase}, \bibinfo{person}{Georgios Smyrnis}, \bibinfo{person}{Thao Nguyen}, \bibinfo{person}{Ryan Marten}, \bibinfo{person}{Mitchell Wortsman}, \bibinfo{person}{Dhruba Ghosh}, \bibinfo{person}{Jieyu Zhang}, {et~al\mbox{.}}} \bibinfo{year}{2023}\natexlab{}.
\newblock \showarticletitle{Datacomp: In search of the next generation of multimodal datasets}.
\newblock \bibinfo{journal}{\emph{Advances in Neural Information Processing Systems}}  \bibinfo{volume}{36} (\bibinfo{year}{2023}), \bibinfo{pages}{27092--27112}.
\newblock


\bibitem[Garber and Djouadi(1988)]%
        {garber1988bounds}
\bibfield{author}{\bibinfo{person}{Frederick~D Garber} {and} \bibinfo{person}{Abdelhamid Djouadi}.} \bibinfo{year}{1988}\natexlab{}.
\newblock \showarticletitle{Bounds on the Bayes classification error based on pairwise risk functions}.
\newblock \bibinfo{journal}{\emph{IEEE Transactions on Pattern Analysis and Machine Intelligence}} \bibinfo{volume}{10}, \bibinfo{number}{2} (\bibinfo{year}{1988}), \bibinfo{pages}{281--288}.
\newblock


\bibitem[Ge et~al\mbox{.}(2024)]%
        {ge2024worldgpt}
\bibfield{author}{\bibinfo{person}{Zhiqi Ge}, \bibinfo{person}{Hongzhe Huang}, \bibinfo{person}{Mingze Zhou}, \bibinfo{person}{Juncheng Li}, \bibinfo{person}{Guoming Wang}, \bibinfo{person}{Siliang Tang}, {and} \bibinfo{person}{Yueting Zhuang}.} \bibinfo{year}{2024}\natexlab{}.
\newblock \showarticletitle{Worldgpt: Empowering llm as multimodal world model}. In \bibinfo{booktitle}{\emph{Proceedings of the 32nd ACM International Conference on Multimedia}}. \bibinfo{pages}{7346--7355}.
\newblock


\bibitem[Geirhos et~al\mbox{.}(2020)]%
        {geirhos2020shortcut}
\bibfield{author}{\bibinfo{person}{Robert Geirhos}, \bibinfo{person}{J{\"o}rn-Henrik Jacobsen}, \bibinfo{person}{Claudio Michaelis}, \bibinfo{person}{Richard Zemel}, \bibinfo{person}{Wieland Brendel}, \bibinfo{person}{Matthias Bethge}, {and} \bibinfo{person}{Felix~A Wichmann}.} \bibinfo{year}{2020}\natexlab{}.
\newblock \showarticletitle{Shortcut learning in deep neural networks}.
\newblock \bibinfo{journal}{\emph{Nature Machine Intelligence}} \bibinfo{volume}{2}, \bibinfo{number}{11} (\bibinfo{year}{2020}), \bibinfo{pages}{665--673}.
\newblock


\bibitem[Gupta et~al\mbox{.}(2019)]%
        {gupta2019lvis}
\bibfield{author}{\bibinfo{person}{Agrim Gupta}, \bibinfo{person}{Piotr Dollar}, {and} \bibinfo{person}{Ross Girshick}.} \bibinfo{year}{2019}\natexlab{}.
\newblock \showarticletitle{Lvis: A dataset for large vocabulary instance segmentation}. In \bibinfo{booktitle}{\emph{Proceedings of the IEEE/CVF conference on computer vision and pattern recognition}}. \bibinfo{pages}{5356--5364}.
\newblock


\bibitem[Ha et~al\mbox{.}(2024)]%
        {ha2024organic}
\bibfield{author}{\bibinfo{person}{Anna Yoo~Jeong Ha}, \bibinfo{person}{Josephine Passananti}, \bibinfo{person}{Ronik Bhaskar}, \bibinfo{person}{Shawn Shan}, \bibinfo{person}{Reid Southen}, \bibinfo{person}{Haitao Zheng}, {and} \bibinfo{person}{Ben~Y Zhao}.} \bibinfo{year}{2024}\natexlab{}.
\newblock \showarticletitle{Organic or diffused: Can we distinguish human art from ai-generated images?}. In \bibinfo{booktitle}{\emph{Proceedings of the 2024 on ACM SIGSAC Conference on Computer and Communications Security}}. \bibinfo{pages}{4822--4836}.
\newblock


\bibitem[Hastie et~al\mbox{.}(2009)]%
        {hastie2009elements}
\bibfield{author}{\bibinfo{person}{Trevor Hastie}, \bibinfo{person}{Robert Tibshirani}, {and} \bibinfo{person}{Jerome Friedman}.} \bibinfo{year}{2009}\natexlab{}.
\newblock \bibinfo{booktitle}{\emph{The Elements of Statistical Learning: Data Mining, Inference, and Prediction} (\bibinfo{edition}{2nd} ed.)}.
\newblock \bibinfo{publisher}{Springer}, \bibinfo{address}{New York, NY}.
\newblock
\showISBNx{978-0-387-84857-0}
\urldef\tempurl%
\url{https://link.springer.com/content/pdf/10.1007/978-0-387-84858-7.pdf}
\showURL{%
\tempurl}


\bibitem[Hu et~al\mbox{.}(2022)]%
        {hu2022lora}
\bibfield{author}{\bibinfo{person}{Edward~J Hu}, \bibinfo{person}{yelong shen}, \bibinfo{person}{Phillip Wallis}, \bibinfo{person}{Zeyuan Allen-Zhu}, \bibinfo{person}{Yuanzhi Li}, \bibinfo{person}{Shean Wang}, \bibinfo{person}{Lu Wang}, {and} \bibinfo{person}{Weizhu Chen}.} \bibinfo{year}{2022}\natexlab{}.
\newblock \showarticletitle{Lo{RA}: Low-Rank Adaptation of Large Language Models}. In \bibinfo{booktitle}{\emph{International Conference on Learning Representations}}.
\newblock
\urldef\tempurl%
\url{https://openreview.net/forum?id=nZeVKeeFYf9}
\showURL{%
\tempurl}


\bibitem[Hu et~al\mbox{.}(2025)]%
        {hu2025membership}
\bibfield{author}{\bibinfo{person}{Yuke Hu}, \bibinfo{person}{Zheng Li}, \bibinfo{person}{Zhihao Liu}, \bibinfo{person}{Yang Zhang}, \bibinfo{person}{Zhan Qin}, \bibinfo{person}{Kui Ren}, {and} \bibinfo{person}{Chun Chen}.} \bibinfo{year}{2025}\natexlab{}.
\newblock \showarticletitle{Membership Inference Attacks Against Vision-Language Models}. In \bibinfo{booktitle}{\emph{Proceedings of the 34th USENIX Security Symposium}}. \bibinfo{publisher}{USENIX Association}.
\newblock


\bibitem[Jagielski et~al\mbox{.}(2023)]%
        {jagielskimeasuring}
\bibfield{author}{\bibinfo{person}{Matthew Jagielski}, \bibinfo{person}{Om Thakkar}, \bibinfo{person}{Florian Tramer}, \bibinfo{person}{Daphne Ippolito}, \bibinfo{person}{Katherine Lee}, \bibinfo{person}{Nicholas Carlini}, \bibinfo{person}{Eric Wallace}, \bibinfo{person}{Shuang Song}, \bibinfo{person}{Abhradeep~Guha Thakurta}, \bibinfo{person}{Nicolas Papernot}, {and} \bibinfo{person}{Chiyuan Zhang}.} \bibinfo{year}{2023}\natexlab{}.
\newblock \showarticletitle{Measuring Forgetting of Memorized Training Examples}. In \bibinfo{booktitle}{\emph{The Eleventh International Conference on Learning Representations}}.
\newblock
\urldef\tempurl%
\url{https://openreview.net/forum?id=7bJizxLKrR}
\showURL{%
\tempurl}


\bibitem[Jia et~al\mbox{.}(2021)]%
        {jia2021scaling}
\bibfield{author}{\bibinfo{person}{Chao Jia}, \bibinfo{person}{Yinfei Yang}, \bibinfo{person}{Ye Xia}, \bibinfo{person}{Yi-Ting Chen}, \bibinfo{person}{Zarana Parekh}, \bibinfo{person}{Hieu Pham}, \bibinfo{person}{Quoc Le}, \bibinfo{person}{Yun-Hsuan Sung}, \bibinfo{person}{Zhen Li}, {and} \bibinfo{person}{Tom Duerig}.} \bibinfo{year}{2021}\natexlab{}.
\newblock \showarticletitle{Scaling up visual and vision-language representation learning with noisy text supervision}. In \bibinfo{booktitle}{\emph{International conference on machine learning}}. PMLR, \bibinfo{pages}{4904--4916}.
\newblock


\bibitem[Kageyama(2025)]%
        {kageyama2025ghibli}
\bibfield{author}{\bibinfo{person}{Yuri Kageyama}.} \bibinfo{year}{2025}\natexlab{}.
\newblock \bibinfo{title}{ChatGPT's viral Studio Ghibli-style images highlight AI copyright concerns}.
\newblock
\urldef\tempurl%
\url{https://apnews.com/article/studio-ghibli-chatgpt-images-hayao-miyazaki-openai-0f4cb487ec3042dd5b43ad47879b91f4}
\showURL{%
\tempurl}


\bibitem[Kim et~al\mbox{.}(2023)]%
        {kim2023propile}
\bibfield{author}{\bibinfo{person}{Siwon Kim}, \bibinfo{person}{Sangdoo Yun}, \bibinfo{person}{Hwaran Lee}, \bibinfo{person}{Martin Gubri}, \bibinfo{person}{Sungroh Yoon}, {and} \bibinfo{person}{Seong~Joon Oh}.} \bibinfo{year}{2023}\natexlab{}.
\newblock \showarticletitle{Propile: Probing privacy leakage in large language models}.
\newblock \bibinfo{journal}{\emph{Advances in Neural Information Processing Systems}}  \bibinfo{volume}{36} (\bibinfo{year}{2023}), \bibinfo{pages}{20750--20762}.
\newblock


\bibitem[Li et~al\mbox{.}(2025)]%
        {li2025llava}
\bibfield{author}{\bibinfo{person}{Bo Li}, \bibinfo{person}{Yuanhan Zhang}, \bibinfo{person}{Dong Guo}, \bibinfo{person}{Renrui Zhang}, \bibinfo{person}{Feng Li}, \bibinfo{person}{Hao Zhang}, \bibinfo{person}{Kaichen Zhang}, \bibinfo{person}{Peiyuan Zhang}, \bibinfo{person}{Yanwei Li}, \bibinfo{person}{Ziwei Liu}, {and} \bibinfo{person}{Chunyuan Li}.} \bibinfo{year}{2025}\natexlab{}.
\newblock \showarticletitle{LLaVA-OneVision: Easy Visual Task Transfer}.
\newblock \bibinfo{journal}{\emph{Transactions on Machine Learning Research}} (\bibinfo{year}{2025}).
\newblock


\bibitem[Li et~al\mbox{.}(2023)]%
        {li2023llava}
\bibfield{author}{\bibinfo{person}{Chunyuan Li}, \bibinfo{person}{Cliff Wong}, \bibinfo{person}{Sheng Zhang}, \bibinfo{person}{Naoto Usuyama}, \bibinfo{person}{Haotian Liu}, \bibinfo{person}{Jianwei Yang}, \bibinfo{person}{Tristan Naumann}, \bibinfo{person}{Hoifung Poon}, {and} \bibinfo{person}{Jianfeng Gao}.} \bibinfo{year}{2023}\natexlab{}.
\newblock \showarticletitle{Llava-med: Training a large language-and-vision assistant for biomedicine in one day}.
\newblock \bibinfo{journal}{\emph{Advances in Neural Information Processing Systems}}  \bibinfo{volume}{36} (\bibinfo{year}{2023}), \bibinfo{pages}{28541--28564}.
\newblock


\bibitem[Li et~al\mbox{.}(2024)]%
        {li2024membership}
\bibfield{author}{\bibinfo{person}{Zhan Li}, \bibinfo{person}{Yongtao Wu}, \bibinfo{person}{Yihang Chen}, \bibinfo{person}{Francesco Tonin}, \bibinfo{person}{Elias Abad~Rocamora}, {and} \bibinfo{person}{Volkan Cevher}.} \bibinfo{year}{2024}\natexlab{}.
\newblock \showarticletitle{Membership inference attacks against large vision-language models}.
\newblock \bibinfo{journal}{\emph{Advances in Neural Information Processing Systems}}  \bibinfo{volume}{37} (\bibinfo{year}{2024}), \bibinfo{pages}{98645--98674}.
\newblock


\bibitem[Lin et~al\mbox{.}(2014)]%
        {lin2014microsoft}
\bibfield{author}{\bibinfo{person}{Tsung-Yi Lin}, \bibinfo{person}{Michael Maire}, \bibinfo{person}{Serge Belongie}, \bibinfo{person}{James Hays}, \bibinfo{person}{Pietro Perona}, \bibinfo{person}{Deva Ramanan}, \bibinfo{person}{Piotr Doll{\'a}r}, {and} \bibinfo{person}{C~Lawrence Zitnick}.} \bibinfo{year}{2014}\natexlab{}.
\newblock \showarticletitle{Microsoft coco: Common objects in context}. In \bibinfo{booktitle}{\emph{Computer vision--ECCV 2014: 13th European conference, zurich, Switzerland, September 6-12, 2014, proceedings, part v 13}}. Springer, \bibinfo{pages}{740--755}.
\newblock


\bibitem[Liu et~al\mbox{.}(2024a)]%
        {liu2024improved}
\bibfield{author}{\bibinfo{person}{Haotian Liu}, \bibinfo{person}{Chunyuan Li}, \bibinfo{person}{Yuheng Li}, {and} \bibinfo{person}{Yong~Jae Lee}.} \bibinfo{year}{2024}\natexlab{a}.
\newblock \showarticletitle{Improved baselines with visual instruction tuning}. In \bibinfo{booktitle}{\emph{Proceedings of the IEEE/CVF Conference on Computer Vision and Pattern Recognition}}. \bibinfo{pages}{26296--26306}.
\newblock


\bibitem[Liu et~al\mbox{.}(2023)]%
        {liu2023visual}
\bibfield{author}{\bibinfo{person}{Haotian Liu}, \bibinfo{person}{Chunyuan Li}, \bibinfo{person}{Qingyang Wu}, {and} \bibinfo{person}{Yong~Jae Lee}.} \bibinfo{year}{2023}\natexlab{}.
\newblock \showarticletitle{Visual instruction tuning}.
\newblock \bibinfo{journal}{\emph{Advances in neural information processing systems}}  \bibinfo{volume}{36} (\bibinfo{year}{2023}), \bibinfo{pages}{34892--34916}.
\newblock


\bibitem[Liu and He(2025)]%
        {liu2025decade}
\bibfield{author}{\bibinfo{person}{Zhuang Liu} {and} \bibinfo{person}{Kaiming He}.} \bibinfo{year}{2025}\natexlab{}.
\newblock \showarticletitle{A Decade's Battle on Dataset Bias: Are We There Yet?}. In \bibinfo{booktitle}{\emph{Proceedings of the International Conference on Learning Representations (ICLR)}}.
\newblock


\bibitem[Liu et~al\mbox{.}(2024b)]%
        {liu2024probing}
\bibfield{author}{\bibinfo{person}{Zhenhua Liu}, \bibinfo{person}{Tong Zhu}, \bibinfo{person}{Chuanyuan Tan}, \bibinfo{person}{Bing Liu}, \bibinfo{person}{Haonan Lu}, {and} \bibinfo{person}{Wenliang Chen}.} \bibinfo{year}{2024}\natexlab{b}.
\newblock \showarticletitle{Probing Language Models for Pre-training Data Detection}. In \bibinfo{booktitle}{\emph{Proceedings of the 62nd Annual Meeting of the Association for Computational Linguistics (Volume 1: Long Papers)}}. \bibinfo{pages}{1576--1587}.
\newblock


\bibitem[Lopez-Paz and Oquab(2017)]%
        {lopez2017revisiting}
\bibfield{author}{\bibinfo{person}{David Lopez-Paz} {and} \bibinfo{person}{Maxime Oquab}.} \bibinfo{year}{2017}\natexlab{}.
\newblock \showarticletitle{Revisiting Classifier Two-Sample Tests}. In \bibinfo{booktitle}{\emph{International Conference on Learning Representations}}.
\newblock


\bibitem[Lu et~al\mbox{.}(2023)]%
        {lu2023seeing}
\bibfield{author}{\bibinfo{person}{Zeyu Lu}, \bibinfo{person}{Di Huang}, \bibinfo{person}{Lei Bai}, \bibinfo{person}{Jingjing Qu}, \bibinfo{person}{Chengyue Wu}, \bibinfo{person}{Xihui Liu}, {and} \bibinfo{person}{Wanli Ouyang}.} \bibinfo{year}{2023}\natexlab{}.
\newblock \showarticletitle{Seeing is not always believing: Benchmarking human and model perception of ai-generated images}.
\newblock \bibinfo{journal}{\emph{Advances in Neural Information Processing Systems}}  \bibinfo{volume}{36} (\bibinfo{year}{2023}), \bibinfo{pages}{25435--25447}.
\newblock


\bibitem[Maini et~al\mbox{.}(2024)]%
        {maini2024llm}
\bibfield{author}{\bibinfo{person}{Pratyush Maini}, \bibinfo{person}{Hengrui Jia}, \bibinfo{person}{Nicolas Papernot}, {and} \bibinfo{person}{Adam Dziedzic}.} \bibinfo{year}{2024}\natexlab{}.
\newblock \showarticletitle{LLM Dataset Inference: Did you train on my dataset?}
\newblock \bibinfo{journal}{\emph{Advances in Neural Information Processing Systems}}  \bibinfo{volume}{37} (\bibinfo{year}{2024}), \bibinfo{pages}{124069--124092}.
\newblock


\bibitem[Maini et~al\mbox{.}(2021)]%
        {mainidataset}
\bibfield{author}{\bibinfo{person}{Pratyush Maini}, \bibinfo{person}{Mohammad Yaghini}, {and} \bibinfo{person}{Nicolas Papernot}.} \bibinfo{year}{2021}\natexlab{}.
\newblock \showarticletitle{Dataset Inference: Ownership Resolution in Machine Learning}. In \bibinfo{booktitle}{\emph{International Conference on Learning Representations}}.
\newblock
\urldef\tempurl%
\url{https://openreview.net/forum?id=hvdKKV2yt7T}
\showURL{%
\tempurl}


\bibitem[Masry et~al\mbox{.}(2022)]%
        {masry2022chartqa}
\bibfield{author}{\bibinfo{person}{Ahmed Masry}, \bibinfo{person}{Xuan~Long Do}, \bibinfo{person}{Jia~Qing Tan}, \bibinfo{person}{Shafiq Joty}, {and} \bibinfo{person}{Enamul Hoque}.} \bibinfo{year}{2022}\natexlab{}.
\newblock \showarticletitle{ChartQA: A Benchmark for Question Answering about Charts with Visual and Logical Reasoning}. In \bibinfo{booktitle}{\emph{Findings of the Association for Computational Linguistics: ACL 2022}}. \bibinfo{pages}{2263--2279}.
\newblock


\bibitem[Mathew et~al\mbox{.}(2021)]%
        {mathew2021docvqa}
\bibfield{author}{\bibinfo{person}{Minesh Mathew}, \bibinfo{person}{Dimosthenis Karatzas}, {and} \bibinfo{person}{CV Jawahar}.} \bibinfo{year}{2021}\natexlab{}.
\newblock \showarticletitle{Docvqa: A dataset for vqa on document images}. In \bibinfo{booktitle}{\emph{Proceedings of the IEEE/CVF winter conference on applications of computer vision}}. \bibinfo{pages}{2200--2209}.
\newblock


\bibitem[Mon-Williams et~al\mbox{.}(2025)]%
        {mon2025embodied}
\bibfield{author}{\bibinfo{person}{Ruaridh Mon-Williams}, \bibinfo{person}{Gen Li}, \bibinfo{person}{Ran Long}, \bibinfo{person}{Wenqian Du}, {and} \bibinfo{person}{Christopher~G Lucas}.} \bibinfo{year}{2025}\natexlab{}.
\newblock \showarticletitle{Embodied large language models enable robots to complete complex tasks in unpredictable environments}.
\newblock \bibinfo{journal}{\emph{Nature Machine Intelligence}} (\bibinfo{year}{2025}), \bibinfo{pages}{1--10}.
\newblock


\bibitem[Montesuma et~al\mbox{.}(2025)]%
        {montesuma2024recent}
\bibfield{author}{\bibinfo{person}{Eduardo~Fernandes Montesuma}, \bibinfo{person}{Fred Maurice~Ngolè Mboula}, {and} \bibinfo{person}{Antoine Souloumiac}.} \bibinfo{year}{2025}\natexlab{}.
\newblock \showarticletitle{Recent Advances in Optimal Transport for Machine Learning}.
\newblock \bibinfo{journal}{\emph{IEEE Transactions on Pattern Analysis and Machine Intelligence}} \bibinfo{volume}{47}, \bibinfo{number}{2} (\bibinfo{year}{2025}), \bibinfo{pages}{1161--1180}.
\newblock
\href{https://doi.org/10.1109/TPAMI.2024.3489030}{doi:\nolinkurl{10.1109/TPAMI.2024.3489030}}


\bibitem[Muennighoff et~al\mbox{.}(2023)]%
        {muennighoff2023scaling}
\bibfield{author}{\bibinfo{person}{Niklas Muennighoff}, \bibinfo{person}{Alexander Rush}, \bibinfo{person}{Boaz Barak}, \bibinfo{person}{Teven Le~Scao}, \bibinfo{person}{Nouamane Tazi}, \bibinfo{person}{Aleksandra Piktus}, \bibinfo{person}{Sampo Pyysalo}, \bibinfo{person}{Thomas Wolf}, {and} \bibinfo{person}{Colin~A Raffel}.} \bibinfo{year}{2023}\natexlab{}.
\newblock \showarticletitle{Scaling data-constrained language models}.
\newblock \bibinfo{journal}{\emph{Advances in Neural Information Processing Systems}}  \bibinfo{volume}{36} (\bibinfo{year}{2023}), \bibinfo{pages}{50358--50376}.
\newblock


\bibitem[Nasr et~al\mbox{.}(2025)]%
        {nasr2025scalable}
\bibfield{author}{\bibinfo{person}{Milad Nasr}, \bibinfo{person}{Javier Rando}, \bibinfo{person}{Nicholas Carlini}, \bibinfo{person}{Jonathan Hayase}, \bibinfo{person}{Matthew Jagielski}, \bibinfo{person}{A.~Feder Cooper}, \bibinfo{person}{Daphne Ippolito}, \bibinfo{person}{Christopher~A. Choquette-Choo}, \bibinfo{person}{Florian Tram{\`e}r}, {and} \bibinfo{person}{Katherine Lee}.} \bibinfo{year}{2025}\natexlab{}.
\newblock \showarticletitle{Scalable Extraction of Training Data from Aligned, Production Language Models}. In \bibinfo{booktitle}{\emph{The Thirteenth International Conference on Learning Representations}}.
\newblock
\urldef\tempurl%
\url{https://openreview.net/forum?id=vjel3nWP2a}
\showURL{%
\tempurl}


\bibitem[{OpenAI}(2021)]%
        {dalle_openai}
\bibfield{author}{\bibinfo{person}{{OpenAI}}.} \bibinfo{year}{2021}\natexlab{}.
\newblock \bibinfo{title}{{DALL·E}: Creating Images from Text}.
\newblock \bibinfo{howpublished}{\url{https://openai.com/dall-e}}.
\newblock


\bibitem[{OpenAI}(2022)]%
        {openai2022chatgpt}
\bibfield{author}{\bibinfo{person}{{OpenAI}}.} \bibinfo{year}{2022}\natexlab{}.
\newblock \bibinfo{title}{Introducing ChatGPT}.
\newblock
\urldef\tempurl%
\url{https://openai.com/blog/chatgpt/}
\showURL{%
\tempurl}


\bibitem[{OpenAI}(2023)]%
        {openai2023gpt4v}
\bibfield{author}{\bibinfo{person}{{OpenAI}}.} \bibinfo{year}{2023}\natexlab{}.
\newblock \bibinfo{title}{GPT-4V(ision) System Card}.
\newblock
\urldef\tempurl%
\url{https://cdn.openai.com/papers/GPTV_System_Card.pdf}
\showURL{%
\tempurl}


\bibitem[OpenAI(2024)]%
        {openai2024gpt4o}
\bibfield{author}{\bibinfo{person}{OpenAI}.} \bibinfo{year}{2024}\natexlab{}.
\newblock \bibinfo{title}{GPT-4o System Card}.
\newblock
\newblock
\shownote{arXiv preprint arXiv:2410.21276}.


\bibitem[Oren et~al\mbox{.}(2024)]%
        {oren2023proving}
\bibfield{author}{\bibinfo{person}{Yonatan Oren}, \bibinfo{person}{Nicole Meister}, \bibinfo{person}{Niladri~S. Chatterji}, \bibinfo{person}{Faisal Ladhak}, {and} \bibinfo{person}{Tatsunori Hashimoto}.} \bibinfo{year}{2024}\natexlab{}.
\newblock \showarticletitle{Proving Test Set Contamination in Black-Box Language Models}. In \bibinfo{booktitle}{\emph{The Twelfth International Conference on Learning Representations}}.
\newblock
\urldef\tempurl%
\url{https://openreview.net/forum?id=KS8mIvetg2}
\showURL{%
\tempurl}


\bibitem[Robison(2025)]%
        {robison2025meta}
\bibfield{author}{\bibinfo{person}{Kylie Robison}.} \bibinfo{year}{2025}\natexlab{}.
\newblock \bibinfo{title}{Meta got caught gaming AI benchmarks}.
\newblock
\urldef\tempurl%
\url{https://www.theverge.com/meta/645012/meta-llama-4-maverick-benchmarks-gaming}
\showURL{%
\tempurl}


\bibitem[Schuhmann et~al\mbox{.}(2022)]%
        {schuhmann2022laion}
\bibfield{author}{\bibinfo{person}{Christoph Schuhmann}, \bibinfo{person}{Romain Beaumont}, \bibinfo{person}{Richard Vencu}, \bibinfo{person}{Cade Gordon}, \bibinfo{person}{Ross Wightman}, \bibinfo{person}{Mehdi Cherti}, \bibinfo{person}{Theo Coombes}, \bibinfo{person}{Aarush Katta}, \bibinfo{person}{Clayton Mullis}, \bibinfo{person}{Mitchell Wortsman}, {et~al\mbox{.}}} \bibinfo{year}{2022}\natexlab{}.
\newblock \showarticletitle{Laion-5b: An open large-scale dataset for training next generation image-text models}.
\newblock \bibinfo{journal}{\emph{Advances in neural information processing systems}}  \bibinfo{volume}{35} (\bibinfo{year}{2022}), \bibinfo{pages}{25278--25294}.
\newblock


\bibitem[Shi et~al\mbox{.}(2024)]%
        {shidetecting}
\bibfield{author}{\bibinfo{person}{Weijia Shi}, \bibinfo{person}{Anirudh Ajith}, \bibinfo{person}{Mengzhou Xia}, \bibinfo{person}{Yangsibo Huang}, \bibinfo{person}{Daogao Liu}, \bibinfo{person}{Terra Blevins}, \bibinfo{person}{Danqi Chen}, {and} \bibinfo{person}{Luke Zettlemoyer}.} \bibinfo{year}{2024}\natexlab{}.
\newblock \showarticletitle{Detecting Pretraining Data from Large Language Models}. In \bibinfo{booktitle}{\emph{The Twelfth International Conference on Learning Representations}}.
\newblock
\urldef\tempurl%
\url{https://openreview.net/forum?id=zWqr3MQuNs}
\showURL{%
\tempurl}


\bibitem[Shokri et~al\mbox{.}(2017)]%
        {shokri2017membership}
\bibfield{author}{\bibinfo{person}{Reza Shokri}, \bibinfo{person}{Marco Stronati}, \bibinfo{person}{Congzheng Song}, {and} \bibinfo{person}{Vitaly Shmatikov}.} \bibinfo{year}{2017}\natexlab{}.
\newblock \showarticletitle{Membership inference attacks against machine learning models}. In \bibinfo{booktitle}{\emph{2017 IEEE symposium on security and privacy (SP)}}. IEEE, \bibinfo{pages}{3--18}.
\newblock


\bibitem[Tan and Le(2019)]%
        {tan2019efficientnet}
\bibfield{author}{\bibinfo{person}{Mingxing Tan} {and} \bibinfo{person}{Quoc Le}.} \bibinfo{year}{2019}\natexlab{}.
\newblock \showarticletitle{Efficientnet: Rethinking model scaling for convolutional neural networks}. In \bibinfo{booktitle}{\emph{International conference on machine learning}}. PMLR, \bibinfo{pages}{6105--6114}.
\newblock


\bibitem[Tirumala et~al\mbox{.}(2022)]%
        {tirumala2022memorization}
\bibfield{author}{\bibinfo{person}{Kushal Tirumala}, \bibinfo{person}{Aram Markosyan}, \bibinfo{person}{Luke Zettlemoyer}, {and} \bibinfo{person}{Armen Aghajanyan}.} \bibinfo{year}{2022}\natexlab{}.
\newblock \showarticletitle{Memorization without overfitting: Analyzing the training dynamics of large language models}.
\newblock \bibinfo{journal}{\emph{Advances in Neural Information Processing Systems}}  \bibinfo{volume}{35} (\bibinfo{year}{2022}), \bibinfo{pages}{38274--38290}.
\newblock


\bibitem[Tong et~al\mbox{.}(2024)]%
        {tong2024cambrian}
\bibfield{author}{\bibinfo{person}{Peter Tong}, \bibinfo{person}{Ellis Brown}, \bibinfo{person}{Penghao Wu}, \bibinfo{person}{Sanghyun Woo}, \bibinfo{person}{Adithya Jairam~Vedagiri IYER}, \bibinfo{person}{Sai~Charitha Akula}, \bibinfo{person}{Shusheng Yang}, \bibinfo{person}{Jihan Yang}, \bibinfo{person}{Manoj Middepogu}, \bibinfo{person}{Ziteng Wang}, {et~al\mbox{.}}} \bibinfo{year}{2024}\natexlab{}.
\newblock \showarticletitle{Cambrian-1: A fully open, vision-centric exploration of multimodal llms}.
\newblock \bibinfo{journal}{\emph{Advances in Neural Information Processing Systems}}  \bibinfo{volume}{37} (\bibinfo{year}{2024}), \bibinfo{pages}{87310--87356}.
\newblock


\bibitem[Ultralytics(2024)]%
        {ultralytics2024yolov11}
\bibfield{author}{\bibinfo{person}{Ultralytics}.} \bibinfo{year}{2024}\natexlab{}.
\newblock \bibinfo{title}{YOLOv11: Real-Time Object Detection}.
\newblock \bibinfo{howpublished}{\url{https://github.com/ultralytics/ultralytics}}.
\newblock


\bibitem[Yeom et~al\mbox{.}(2018)]%
        {yeom2018privacy}
\bibfield{author}{\bibinfo{person}{Samuel Yeom}, \bibinfo{person}{Irene Giacomelli}, \bibinfo{person}{Matt Fredrikson}, {and} \bibinfo{person}{Somesh Jha}.} \bibinfo{year}{2018}\natexlab{}.
\newblock \showarticletitle{Privacy risk in machine learning: Analyzing the connection to overfitting}. In \bibinfo{booktitle}{\emph{2018 IEEE 31st computer security foundations symposium (CSF)}}. IEEE, \bibinfo{pages}{268--282}.
\newblock


\bibitem[Yu et~al\mbox{.}(2016)]%
        {yu2016modeling}
\bibfield{author}{\bibinfo{person}{Licheng Yu}, \bibinfo{person}{Patrick Poirson}, \bibinfo{person}{Shan Yang}, \bibinfo{person}{Alexander~C Berg}, {and} \bibinfo{person}{Tamara~L Berg}.} \bibinfo{year}{2016}\natexlab{}.
\newblock \showarticletitle{Modeling context in referring expressions}. In \bibinfo{booktitle}{\emph{Computer Vision--ECCV 2016: 14th European Conference, Amsterdam, The Netherlands, October 11-14, 2016, Proceedings, Part II 14}}. Springer, \bibinfo{pages}{69--85}.
\newblock


\bibitem[Zeng et~al\mbox{.}({[n.\,d.]})]%
        {zengunderstanding}
\bibfield{author}{\bibinfo{person}{Boya Zeng}, \bibinfo{person}{Yida Yin}, {and} \bibinfo{person}{Zhuang Liu}.} \bibinfo{year}{[n.\,d.]}\natexlab{}.
\newblock \showarticletitle{Understanding Bias in Large-Scale Visual Datasets}. In \bibinfo{booktitle}{\emph{The Thirty-eighth Annual Conference on Neural Information Processing Systems}}.
\newblock


\bibitem[Zhang et~al\mbox{.}(2025)]%
        {zhang2025min}
\bibfield{author}{\bibinfo{person}{Jingyang Zhang}, \bibinfo{person}{Jingwei Sun}, \bibinfo{person}{Eric Yeats}, \bibinfo{person}{Yang Ouyang}, \bibinfo{person}{Martin Kuo}, \bibinfo{person}{Jianyi Zhang}, \bibinfo{person}{Hao~Frank Yang}, {and} \bibinfo{person}{Hai Li}.} \bibinfo{year}{2025}\natexlab{}.
\newblock \showarticletitle{Min-K\%++: Improved Baseline for Pre-Training Data Detection from Large Language Models}. In \bibinfo{booktitle}{\emph{International Conference on Learning Representations (ICLR)}}.
\newblock


\bibitem[Zhu et~al\mbox{.}(2024)]%
        {zhu2024reliable}
\bibfield{author}{\bibinfo{person}{Hongyu Zhu}, \bibinfo{person}{Sichu Liang}, \bibinfo{person}{Wentao Hu}, \bibinfo{person}{Li Fangqi}, \bibinfo{person}{Ju Jia}, {and} \bibinfo{person}{Shi-Lin Wang}.} \bibinfo{year}{2024}\natexlab{}.
\newblock \showarticletitle{Reliable Model Watermarking: Defending Against Theft without Compromising on Evasion}. In \bibinfo{booktitle}{\emph{Proceedings of the 32nd ACM International Conference on Multimedia}}. \bibinfo{pages}{10124--10133}.
\newblock


\bibitem[Zhu et~al\mbox{.}(2003)]%
        {zhu2003semi}
\bibfield{author}{\bibinfo{person}{Xiaojin Zhu}, \bibinfo{person}{Zoubin Ghahramani}, {and} \bibinfo{person}{John~D Lafferty}.} \bibinfo{year}{2003}\natexlab{}.
\newblock \showarticletitle{Semi-supervised learning using gaussian fields and harmonic functions}. In \bibinfo{booktitle}{\emph{Proceedings of the 20th International conference on Machine learning (ICML-03)}}. \bibinfo{pages}{912--919}.
\newblock


\end{thebibliography}

\end{document}


\title{Supplementary Material of Revisiting Data Auditing in Large Vision-Language Models}

\maketitle

\section*{Overview}
The complete experimental results, detailed hyperparameter settings, and the code used to reproduce them are available in our \textbf{anonymous repository}: \url{https://anonymous.4open.science/r/Revisiting-VLM-MIA-EB70}.

As outlined in the main text, we evaluate our method against baselines such as \textbf{Min-K\%} with \( K = 5 \), as well as with varying \( K \) values ranging from 0 to 90 in increments of 10. While the main paper reports the best results achieved by each baseline across these different \( K \) values, the anonymous repository provides a comprehensive overview of the full performance variation across \( K \) for further research and analysis.

Similarly, for the \textbf{MaxRényi-K\%} baseline, where different values of the regularization coefficient are considered, we collect and report the optimal performing results.

A sample of the raw data collected is presented in Table~\ref{tab:sample}.

 For more comprehensive and detailed findings, we invite reviewers to consult the \textbf{anonymous repository}, which contains in-depth analyses and extended discussions. These cover the superior performance of our WiRED method under standard metrics such as FID, as well as concrete examples demonstrating how BER effectively captures the gap between theoretically achievable accuracy and practical performance in real-world classification tasks.

\begin{table}[htbp]
  \centering
  \caption{Sample results of membership inference using LLavA-1.5 on the COCO dataset with descriptions.}
  \resizebox{\linewidth}{!}{
    \begin{tabular}{l|ccc}
      \toprule
      Model & AUC & Accuracy & TPR@5\%FPR \\
      \midrule
      ppl   & 0.5371 & 0.5417 & 0.0167 \\
      Min0\% Prob & 0.5086 & 0.5250 & 0.0133 \\
      Min5.0\% Prob & 0.5086 & 0.5250 & 0.0133 \\
      Min10.0\% Prob & 0.5201 & 0.5317 & 0.0100 \\
      Min20.0\% Prob & 0.5317 & 0.5450 & 0.0133 \\
      Min30.0\% Prob & 0.5384 & 0.5450 & 0.0267 \\
      Min40.0\% Prob & 0.5408 & 0.5533 & 0.0333 \\
      Min50.0\% Prob & 0.5415 & 0.5483 & 0.0300 \\
      Min60.0\% Prob & 0.5409 & 0.5467 & 0.0133 \\
      Min70.0\% Prob & 0.5388 & 0.5417 & 0.0200 \\
      Min80.0\% Prob & 0.5368 & 0.5417 & 0.0167 \\
      Min90.0\% Prob & 0.5367 & 0.5417 & 0.0167 \\
      Modrenyi1 & 0.5373 & 0.5450 & 0.0200 \\
      Modrenyi05 & 0.5369 & 0.5383 & 0.0200 \\
      Modrenyi2 & 0.5370 & 0.5400 & 0.0333 \\
      MaxProbGap & 0.5304 & 0.5450 & 0.0567 \\
      Min0\% Prob++ & 0.5034 & 0.5200 & 0.0567 \\
      Min5.0\% Prob++ & 0.5034 & 0.5200 & 0.0567 \\
      Min10.0\% Prob++ & 0.5286 & 0.5417 & 0.0400 \\
      Min20.0\% Prob++ & 0.5299 & 0.5383 & 0.0767 \\
      Min30.0\% Prob++ & 0.5257 & 0.5317 & 0.0833 \\
      Min40.0\% Prob++ & 0.5210 & 0.5283 & 0.0767 \\
      Min50.0\% Prob++ & 0.5241 & 0.5300 & 0.0733 \\
      Min60.0\% Prob++ & 0.5272 & 0.5350 & 0.0667 \\
      Min70.0\% Prob++ & 0.5310 & 0.5367 & 0.0533 \\
      Min80.0\% Prob++ & 0.5340 & 0.5400 & 0.0700 \\
      Min90.0\% Prob++ & 0.5336 & 0.5367 & 0.0633 \\
      Min100\% Prob++ & 0.5348 & 0.5367 & 0.0400 \\
      \bottomrule
      \multicolumn{4}{l}{\textit{Note: Additional results are available in the anonymous repository.}} \\
    \end{tabular}%
  }
  \label{tab:sample}%
\end{table}%